# Fuzzy Adaptive Teaching Learning-based Optimization Strategy for the Problem of Generating Mixed Strength *t*-way Test Suites

**KAMAL Z. ZAMLI and FAKHRUD DIN**
*IBM Centre of Excellence*
*Faculty of Computer Systems and Software Engineering*
*Universiti Malaysia Pahang*
*Lebuhraya Tun Razak, 26300 Kuantan, Pahang Darul Makmur, Malaysia*
*Email: kamalz@ump.edu.my*

**SALMI BAHAROM**
*Faculty of Computer Science and Information Technology*
*Universiti Putra Malaysia*
*Email: salmi@upm.edu.my*

**BESTOUN S. AHMED**
*Istituto Dalle Molle di Studi sull'Intelligenza Artificiale (IDSIA) Galleria 2, CH-6928 Manno-Lugano, Switzerland*
*Email: bestoun@idsia.ch*

## Abstract

The teaching learning-based optimization (TLBO) algorithm has shown competitive performance in solving numerous real-world optimization problems. Nevertheless, this algorithm requires better control for exploitation and exploration to prevent premature convergence (i.e., trapped in local optima), as well as enhance solution diversity. Thus, this paper proposes a new TLBO variant based on Mamdani fuzzy inference system, called ATLBO, to permit adaptive selection of its global and local search operations. In order to assess its performances, we adopt ATLBO for the mixed strength *t*-way test generation problem. Experimental results reveal that ATLBO exhibits competitive performances against the original TLBO and other meta-heuristic counterparts.

## 1. Introduction

In the past decades, a few meta-heuristic algorithms have been proposed in scientific literature to address real-world optimization problems. These algorithms mainly comprises exploration and exploitation (or diversification and intensification) [1]. Exploration roams the random search space on a global scale (i.e., global search), whereas exploitation focuses on searching in a local region by exploiting the current suitable solution (i.e., local search). Overemphasizing exploration consumes significant computational resources and prevents convergence. Conversely, excessive exploitation tends to deny a diverse solution and may lead toward local optima. Most meta-heuristic algorithms introduce specific parameter controls to manage exploration and exploitation effectively. For example, genetic algorithm (GA) [2] exploits mutation and crossover rate; particle swarm optimization (PSO) [3] introduces inertia weight and social/cognitive parameters; harmony search (HS) [4] relies on the consideration rate of harmony memory and pitch adjustment; and ant colony optimization (ACO) [5] exploits evaporation rate, pheromone influence, and heuristic influence. Tuning the parameters accordingly ensures a suitable quality solution. However, the tuning of these parameters is often time consuming and problem specific because a single size is unavailable to fits all approaches.

The teaching learning-based optimization algorithm (TLBO) [6, 7] adopts a simplistic approach of disregarding the control parameters (i.e., parameter free). TLBO specifically performs both global and local search sequentially per iteration to balance exploration and exploitation. Given that exploration and exploitation are dynamic in nature depending on the current search space region, any preset division between the two can be counter-productive and may lead to poor quality solutions. This paper addresses these issues through a new TLBO variant, adaptive TLBO (ATLBO) integrated with the Mamdani-type fuzzy inference system [8, 9]. ATLBO adaptively selects its local and global search operations. In order to assess its performances, we adopt ATLBO for the mixed strength *t*-way test generation problem.

Our contributions are summarized as follows:

- The novel ATLBO strategy based on the Mamdani-type fuzzy inference system is presented for exploration (i.e., global search) and exploitation (i.e., local search) selection.
- ATLBO is the first TLBO-variant strategy that addresses generation for both uniform and mixed-strength *t*-way test suite.

This study is organized as follows. Section 2 presents the theoretical framework that covers the generation problem of *t*-way test and its mathematical notation. Section 3 describes the related work. Section 4 highlights the original TLBO algorithm and its variants, along with its applications. Section 5 outlines the novel ATLBO. Benchmark experiments are presented in Section 6. Section 7 and 8 discusses the experimental observations and validity threats, respectively. Finally, Section 9 concludes this study and presents the scope for future work.

## 2. Covering Array (CA) and the Generation Problem of Mixed-Strength *t*-way Test

The generation problem of *t*-way test is often associated with CA notation, where $t$ represents the desired interaction strength. A CA ($N$; $t$, $p$, $v$), which is also expressed as CA ($N$; $t$, $v^p$), is a combinatorial structure constructed as an array of $N$ rows and $p$ columns (i.e., parameters) on $v$ values, such that every $N \times t$ sub-array contains all ordered subsets from the $v$ values of size $t$ at least once [10]. When the number of component values varies, this condition can be handled by a mixed CA (MCA) ($N$; $t$, $p$, ($v_1$, $v_2$, ...$v_i$)) or MCA ($N$; $t$, $v_1^{p1}$, $v_2^{p2}$, …$v_i^{pi}$). A mixed-strength CA is defined to address the impact of non-uniform interaction. A mixed-strength CA (or variable-strength CA; VCA) ($N$; $t$; $p$, $v$, ($CA_1…CA_j$)) is also a combinatorial structure constructed as an array of $N$ rows and $p$ column on $v$ values. However, every $N \times t$ array of VCA contains one or more sub-CAs, namely, $CA_1...CA_i$, each of which has an interaction strength $t_1…t_j$ that is larger than $t$.

A simple model of online gaming architecture (Table 1) is utilized to illustrate the generation problem of *t*-way test. The online gaming architecture comprises five parameters, i.e., server, game server, smart phone OS, database server, and client browser. The online gaming architecture depicted can generally be summarized as a system of five parameters with a combination of three parameters with two values (i.e., server = {subscription, trial account}, game server = {dedicated, peer-to-peer}, and smart phone OS = {iOS, Android}) and two parameters with three values (i.e., database server = {Oracle, MySQL, SQL Server} and client browser = {Google Chrome, Opera, Internet Explorer}). The CA notation provided earlier is utilized to express the online gaming architecture as MCA ($N$; $t$, $2^3$ $3^2$).

Table 1. Online Gaming Architecture: Parameters and Values

Main Strength $t = 2$

Sub-strength $t = 3$

| **Server** | **Game Server** | **Smart Phone OS** | **Database Server** | **Client Browser** |
|---|---|---|---|---|
| Subscription | Dedicated | iOS | Oracle | Google Chrome |
| Trial Account | Peer-to-Peer | Android | MySQL | Opera |
| | | | SQL Server | Internet Explorer |

At full interaction strength $t = 5$, the online gaming architecture (with MCA ($N$; 5, $2^3$ $3^2$)) yields exhaustive combination with 72 (2 × 2 × 2 × 3 × 3) test cases. Given the mixed-strength interaction (with main strength $t = 2$ and sub-strength $t = 3$), the online gaming architecture (with VCA ($N$; 2, $2^3$ $3^2$, $CA$ (3, $2^3$))) yields 10 test cases (i.e., a reduction of 86.11% from the 72 exhaustive possibilities). Table 2 highlights the corresponding test case mapping from the abovementioned mixed-strength CAs. The selection of the (mixed-strength) CA representation depends on product requirements and creativity of test engineers based on the given testing problem.

Table 2. Mixed-Strength VCA ($N$; 2, $2^3$ $3^2$, $CA$ (3, $2^3$))

| **Test ID** | **Server** | **Game Server** | **Smart Phone OS** | **Database Server** | **Client Browser** |
|---|---|---|---|---|---|
| 1 | Trial Account | Dedicated | iOS | MySQL | Google Chrome |
| 2 | Subscription | Dedicated | Android | SQL Server | Internet Explorer |
| 3 | Subscription | Peer-to-Peer | iOS | Oracle | Opera |
| 4 | Trial Account | Peer-to-Peer | Android | Oracle | Google Chrome |
| 5 | Trial Account | Peer-to-Peer | iOS | SQL Server | Internet Explorer |
| 6 | Subscription | Peer-to-Peer | Android | MySQL | Opera |
| 7 | Trial Account | Dedicated | Android | SQL Server | Opera |
| 8 | Subscription | Dedicated | iOS | SQL Server | Google Chrome |
| 9 | Trial Account | Dedicated | iOS | Oracle | Internet Explorer |
| 10 | Trial Account | Dedicated | iOS | MySQL | Internet Explorer |

the generation of $t$-way test can be mathematically expressed as an optimization problem utilizing Equations 1 and 2 as follows:

$$Maximize\ f(x) = \sum_{1}^{N} x_i \qquad (1)$$

$$\text{Subject to } x \in x_i, i = 1, 2, \ldots, N \qquad (2)$$

where $f(x)$ is an objective function that captures the weight of the test case in terms of the number of covered interactions; $x$ is the set of each decision variable $x_i$; $x_i$ is the set of a possible range of values for each decision variable, i.e., $x_i = \{x_\text{i}\ (1), x_i(2), \ldots, x_i(K)\}$ for discrete decision variables ($x_\text{i}\ (1) < x_i(2) < \ldots < x_i(K)$); $N$ is the number of decision parameters; and $K$ is the number of possible values for the discrete variables.

## 3. Meta-Heuristic-based *t*-way Strategies

The generation of *t*-way test suite is an NP-hard problem [11]. Significant research efforts have been conducted to investigate this *t*-way test suite generation problem. Recent efforts have focused on the adoption of meta-heuristic algorithms as the basis for the *t*-way strategy because these algorithms can achieve better results in terms of CA sizes compared with other computational methods [12].

Meta-heuristic-based strategies often start with a population of random solutions. One or more search operators are then iteratively applied to the population to improve the overall fitness (i.e., in terms of greedily covering the interaction combinations). Although several variations exist, the main difference among meta-heuristic strategies lies on each individual search operator and on the manipulation of exploration and exploitation. In line with the upcoming field called search-based software engineering [13-15], a few newly developed *t*-way strategies based on meta-heuristics have been introduced in literature.

Genetic algorithm (GA), ant colony optimization (ACO), and simulated annealing (SA) represent early attempts in adopting meta-heuristic algorithms for generating *t*-way tests. GA [16] mimics the natural selection processes and begins with randomly created test cases, which are referred to as chromosomes. These chromosomes undergo crossover and mutation until a termination criterion is met. The best chromosomes are (probabilistically) selected and added to the final test suite in each cycle. Unlike GA, ACO [16] mimics the behavior of ants in their food search. SA [17] relies on a large random search space and probability-based transformation equations to generate a *t*-way test suite. GA- and ACO-based strategies have been criticized for their steep learning curve, complex algorithm structure, and potential requirement of large computational resources. SA, being a single solution meta-heuristic, can be overly sensitive to its initial starting point in the search space, hence, prone to suffer from early convergence.

Harmony search strategy (HSS) [18] is a meta-heuristic *t*-way strategy based on the harmony search algorithm . HSS mimics the behavior of musicians who attempt to compose exceptional music either from improvisations (i.e., modifying a tune from their memory) or from random sampling. HSS achieves its mimicry by iteratively exploiting the harmony memory to store the acceptable solution through several defined improvisations within its local and global search processes. One test case is selected in each improvisation to be part of the final test suite until all the required interactions are covered.

Cuckoo search (CS) [19] is a meta-heuristic *t*-way strategy that mimics the unique lifestyle and aggressive reproduction strategy of Cuckoo birds. CS initially generates random initial eggs on another nest of a host bird. Each egg in a nest represents a vector solution (i.e., a test case). Two operations are performed at each generation. A new nest is initially generated (typically through a Levy flight path) and evaluated against the existing nests. The new nest then replaces the current one if it has a better objective function. Subsequently, CS adopts probabilistic elitism to maintain elite solutions for the next generation.

Discrete particle swarm optimization (DPSO) [20], particle swarm test generator (PSTG) [21-24], and adaptive particle swarm optimization (APSO) [12] are meta-heuristic-based *t*-way strategies based on several variants of particle swarm optimization. PSO-based *t*-way strategies mimic the swarm behavior of flocking birds to perform a search. The global and local searches within PSO are guided through its inertia weight and social/cognitive parameters. A random swarm is initially created. Thereafter, PSO iteratively selects the candidate solution within the swarm to be added to the final suite until all the interaction tuples are covered. Unlike DPSO and PSTG, APSO does not require tuning, because its control parameters (i.e., inertia weight and social/cognitive parameters) are dynamically calibrated utilizing the Mamdani-type fuzzy inference system. Similar to APSO, the proposed ATLBO in this study also adopts the Mamdani-type fuzzy inference system. Unlike APSO, the Mamdani-type fuzzy inference system is not utilized for automatic parameter tuning in this study; instead, the fuzzy interference system is applied as the selection mechanism for the global and local search operations.

## 4. Original TLBO and Its Variants

Given its simplicity, TLBO has been actively adopted to solve optimization problems in many application areas of science and engineering, such as in [25-28]. TLBO [6, 7] basically takes an analogy from the teaching and learning process between a teacher and his students. A teacher is assumed to be more knowledgeable than his students (i.e., with better fitness value). The teacher imparts his knowledge to his students to match with his competency level. Given that teachers also have different competency levels, potential improvements can occur if students learned from other teachers as well (in any subsequent iterations). Students can also learn from one another simultaneously to improve their competency level.

The solution is represented in the population $X$ within TLBO. An individual $X_i$ within the population represents a single possible solution. $X_i$ is specifically a vector with $D$ elements, where $D$ is the dimension of the problem that represents the subjects taken by the students or taught by the teacher.

TLBO divides the entire searching process into two main phases: teacher and learner. TLBO undergoes both phases sequentially per iteration to perform the search (Figure 1). The teacher phase invokes the global search operation (i.e., exploration). The teacher is always assigned to the best individual $X_i$ at any instance of the search process. The algorithm attempts to improve other individual $X_i$ by moving their position toward $X_{teacher}$ by considering the current mean value of the population $X_{mean}$, as shown as follows:

$$X_i^{t+1} = X_i^t + r(X_{teacher} - T_F X_{mean}) \qquad (3)$$

where $X_i^{t+1}$ is the newly updated $X_i^t$, $X_{teacher}$ is the best individual in the $X$ population, $X_{mean}$ is the mean of the $X$ population, $r$ is a random number from [0, 1], and $T_F$ is a teaching factor that can either be 1 or 2 to emphasize the quality of students.

The learner phase exploits the local search operation (i.e., exploitation). The learner $X_i^t$ specifically increases its knowledge by interacting with its random peer $X_j^t$ within the $X$ population (i.e., $i \neq j$). A learner learns if and only if the other learner has more knowledge than he does. At any iteration $i$, if $X_i^t$ is better than $X_j^t$, then $X_j^t$ moves toward $X_i^t$ (Equation 4). Otherwise, $X_i^t$ moves toward $X_j^t$ (Equation 5).

$$X_i^{t+1} = X_i^t + r\left(X_j^t - X_i^t\right) \qquad (4)$$

$$X_i^{t+1} = X_i^t + r\left(X_i^t - X_j^t\right) \qquad (5)$$

where $X_i^{t+1}$ is the newly updated $X_i^t$, $X_j^t$ is the random peer, and $r$ is a random number from [0, 1].

The original TLBO is summarized in Figure 1.

**Algorithm 1:** The Original TLBO Algorithm

**Input:** the population $X = X_1, X_2 \ldots X_D$
**Output:** $X_{best}$ and the updated population $X' = \{X'_1, X'_2, \ldots, X'_D\}$

Initialize random populations of learners $X$ and evaluate all learners $X$
**while** *stopping criteria not met* **do**
  **for** $i = 1$ *to population size* **do**
    `/* Teacher Phase ... Exploration */`
    Select $X_{teacher}$ and calculate $X_{mean}$
    $T_F = round(1 + r(0,1))$
    $X_i^{t+1} = X_i^t + r(X_{teacher} - T_F X_{mean})$
    **if** $f(X_i^{t+1})$ *is better than* $f(X_i^t)$ **then**
      $X_i^t = X_i^{t+1}$
    `/* Learner Phase ... Exploitation */`
    Randomly select one learner $X_j^t$ from the population X such that $i \neq j$
    **if** $f(X_i^t)$ *is better than* $f(X_j^t)$ **then**
      $X_i^{t+1} = X_i^t + r(X_j^t - X_i^t)$
    **else**
      $X_i^{t+1} = X_i^t + r(X_i^t - X_j^t)$
    **if** $f(X_i^{t+1})$ *is better than* $f(X_i^t)$ **then**
      $[X_i^t = X_i^{t+1}$
  Get best result $X_{best}$

Figure 1. Original TLBO Algorithm

### 4.1 Review of TLBO Variants and Their Applications

Since the inception of TLBO, many of its variants have been introduced to improve its performance. Apart from the original TLBO, the main TLBO variants available in literature can be divided into three categories: modified-, hybrid-, and cooperative-based algorithms. A discussion with several examples of each category is presented below.

The modified-based category refers to variants that enhance the performance of TLBO by modifying its parameter (e.g., elitism feature and adaptive behavior) or altering the teacher and/or learner phases. Rao and Patel [29] introduced the elitism feature within TLBO and demonstrated its efficiency in attempting 35 constrained benchmark functions. Niknam et al. [30] introduced an additional phase called the modified phase, whereby four adaptive search operators are defined and probabilistically selected during runtime. This study has been successfully adopted for dynamic economic dispatch in power system. Based on the work of Niknam et al., Amin et al. [31] also exploited the modified phase within TLBO and introduced an adaptive search operator based on the Morlet wavelet function. The modified TLBO is then adopted for solving the problem of a multi-objective optimal power flow using the fuzzy decision support (i.e., selecting the best Pareto-optimal solution). Although not introducing new phase, Hoseini et al [32] adopted similar approach for addressing multi-objective optimal location of automatic voltage regulators in distribution system. Mandal and Roy [33] solved the problem of a multi-objective optimal reactive power dispatch by incorporating the quasi-opposition-based learning concept in the original TLBO algorithm to accelerate the convergence speed. Xia et al. [34] more recently presented a modified TLBO for problem of disassembly sequence planning. They modified the teacher–learner operator apart from introducing a feasible solution generator operator to satisfy the constraints of a disassembly sequence.

The modified-based TLBO algorithm produces sound results; however, it is often applicable to specific problem and not sufficiently general (i.e., because of the problem domain assumption). Thus, the performance of the modified TLBO cannot be guaranteed even with the slight modification of the same instances of problem.

The hybrid-based category refers to the integration of one or more meta-heuristic algorithms (or their search operators) within TLBO, which complements the modified-based category. To date, TLBO has been utilized to form a hybrid model from several meta-heuristic algorithms. Jiang and Zhou [35] explored the adoption of a hybrid TLBO with differential evolution to solve the short-term optimal hydro-thermal scheduling. Tuo et al. [36] implemented an improved HS-based TLBO to balance the convergence speed and population diversity for the general constrained optimization problem. Lim and Mat-Isa [37] integrated PSO with TLBO as an alternative strategy for local optimum problem within the constrained benchmark functions. Huang et al. [38] recently integrated TLBO with the CS algorithm for parameter optimization in structure designing and machining problem.

Although hybrid-based algorithm can be useful for capitalizing on TLBO strengths and compensating its deficiencies, the actual implementation can be bulky and computationally heavy. Moreover, achieving a suitable balance between exploration and exploitation (i.e., of the hybrid search operators) can still be problematic.

Finally, the cooperative-based category refers to TLBO variants that address large optimization problem with multiple-swarm populations. Tasks are split in *k* sub-problems for simultaneous optimization before combining the results in this category. Biswas et al. [39] highlighted the earliest work that exploits cooperative co-evolutionary TLBO with a modified exploration strategy for large-scale optimization problem. Similarly, Satapathy and Naik [40] explored cooperative TLBO, which allows cooperative behavior by adopting multiple-swarm populations. Zou et al. [41] proposed the adoption of multiple-swarm populations for dynamic optimization problem.

Despite the potential of cooperative-based TLBO algorithm, its key challenges are twofold: to identify the suitable sub-problem size (and multiple-swarm populations) and to model the independent variables for different sub-problems.

## 5. Proposed Fuzzy ATLBO

The proposed ATLBO is based on the Mamdani-type fuzzy inference system [8, 9], as shown in Figure 2 with three inputs (i.e., quality measure $Q_m$, intensification measure $I_m$, and diversification measure $D_m$) and one output (i.e., selection).

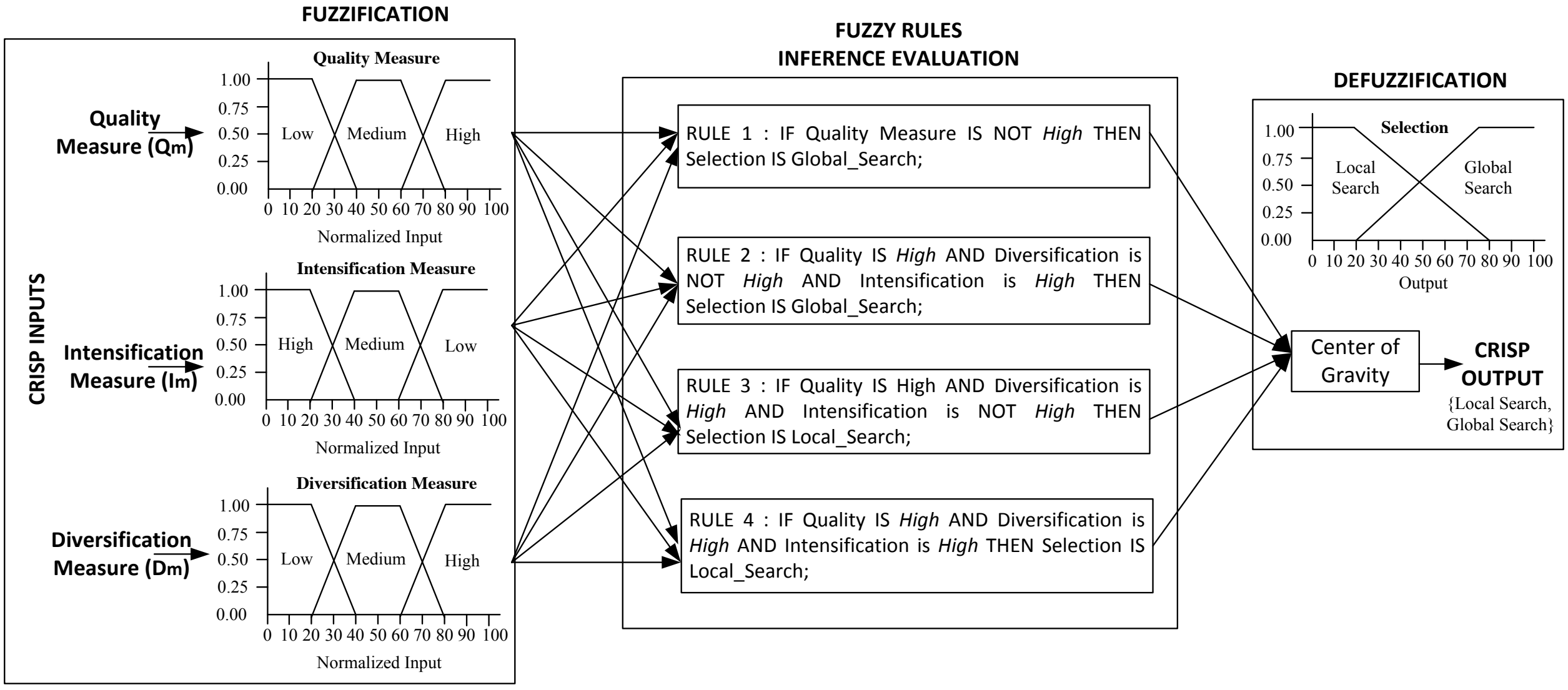

Figure 2. Fuzzy Inference System for ATLBO

Referring to the crisp inputs in Figure 2, the quality measure ($Q_m$) is the normalized fitness value capturing the quality of the current potential solution, $X_{current}$. $Q_m$ can be formally defined as:

$$Q_m = \left[\frac{X_{current\,fitness} - min\,fitness}{max\,fitness - min\,fitness}\right] \cdot 100 \qquad [6]$$

The intensification measure $I_m$ is the normalized value of a Hamming distance that measures the proximity of the $X_{current}$ against $X_{best}$. $I_m$ can be formally defined as follows:

$$I_m = \left[\frac{|X_{best} - X_{current}|}{D}\right] \cdot 100 \qquad [7]$$

where $D$ is the dimension of vector $X$.

The diversification measure $D_m$ is also the normalized value of a Hamming distance. Unlike $I_m$, which measures the intensification of the search against the global best, $D_m$ measures the diversity of $X_{current}$ against the overall $X$ population. $D_m$ can be formally defined as follows:

$$D_m = \left[\frac{\sum_{j=1}^{population\,size} |X_j - X_{current}|}{D}\right] \cdot 100 \qquad [8]$$

The fuzzification process is based on three defined trapezoidal membership functions with linguistic terms, i.e., *low*, *medium*, and *high*. The trapezoidal membership functions for the $Q_m$ and $D_m$ are identical. The values in the range of 0 to 20 are considered as absolute *low*; the values in the range of 20 to 40 are considered as partial *low* and *medium*; the values in the range of 40 to 60 are considered as absolute *medium*; the values in the range of 60 to 80 are considered as partial *medium* and *high*; and the values in the range of 80 to 100 are considered as absolute *high*. The *high* and *low* ranges in the case of $I_m$ are exchanged. Conversely, change did not occur in the *medium* range.

Four defined fuzzy rules (Figure 2) exist in the inference evaluation fuzzy rules based on the following scenarios:

- Rule 1: Quality measure is *low* regardless of intensification and diversification measure. The search is trapped in the local minima region, thus requiring a global search.

- Rule 2: Quality measure is *high* but lacks diversity. The search is trapped in the local minima region because of excessive local search.
- Rule 3: Quality measure is *high* but lacks convergence because of excessive global search.
- Rule 4: Search is near convergence. Local search is required.

A single output called selection is defined for defuzzification. Selection has two linguistic terms called *local_search* and *global_search*, which are represented by the trapezoidal membership function. The selection values in the range of 0 to 20 are considered as absolute *local search*; the selection values in the range of 20 to 80 are considered as partial *local search* and *global search*; and the selection values in the range of 80 to 100 are considered as absolute *global search*. Eventually, the actual selection depends on the output of the defuzzification process based on the center of gravity. The selection is set to *global search* when the defuzzification output is larger than 50%. Otherwise, the selection is set to *local search*.

Figure 3 highlights the newly developed ATLBO based on the defined fuzzy inference system and TLBO description provided in the previous section.

**Algorithm 2:** The General Adaptive TLBO based on Fuzzy Inference System

**Input:** the population $X = X_1, X_2 \ldots X_D$
**Output:** $X_{best}$ and the updated population $X' = \{X'_1, X'_2, \ldots, X'_D\}$

Define of member functions for the linguistic variables
Define fuzzy rules
Initialize random populations of learners $X$ and evaluate the mean of all learners $X_{mean}$
$X_{best}$ = Randomly select a learner from $X$
$X_{best} = X_{teacher}$
**while** *stopping criteria not met* **do**
    **for** $i = 1$ *to population size* **do**
        Compute $Q_m$, $I_m$ and $D_m$
        Fuzzify based on $Q_m$, $I_m$ and $D_m$
        Defuzzify and set Selection = crisp output
        **if** $Selection > 50$ **then**
            /* Teacher Phase ... Exploration */
            Select $X_{teacher}$ and calculate $X_{mean}$
            $T_F = round(1 + r(0,1))$
            $X_i^{t+1} = X_i^t + r(X_{teacher} - T_F X_{mean})$
            **if** $f(X_i^{t+1})$ *is better than* $f(X_i^t)$ **then**
                $X_i^t = X_i^{t+1}$
        **else**
            /* Learner Phase ... Exploitation */
            Randomly select one learner $X_j^t$ from the population X such that $i \neq j$
            **if** $f(X_i^t)$ *is better than* $f(X_j^t)$ **then**
                $X_i^{t+1} = X_i^t + r(X_j^t - X_i^t)$
            **else**
                $X_i^{t+1} = X_i^t + r(X_i^t - X_j^t)$
            **if** $f(X_i^{t+1})$ *is better than* $f(X_i^t)$ **then**
                $[X_i^t = X_i^{t+1}$
    Get best result $X_{best}$

Figure 3. General ATLBO based on the Fuzzy Inference System

## 6. ATLBO for Mixed-Strength *t*-way Test Suite Generation

After providing an overview of TLBO and its adaptive variant (ATLBO), the following section outlines its application to address the problem of generating mixed-strength *t*-way test suite. ATLBO is generally a composition of two main algorithms: (1) an interaction element-generation algorithm, which generates combinations of parameter values that are utilized in the test suite generator for optimization purposes; and (2) an ATLBO-based test suite generator algorithm. The next sub-sections explain these two algorithms in detail.

### 6.1 Interaction Elements Generation Algorithm

The interaction element-generation algorithm involves generating the parameter $P$ combinations and values $v$ for each parameter combination. The parameter generation adopts binary digits, whereby 0 indicates the exclusion of a referred parameter, and 1 indicates the inclusion of the parameter.

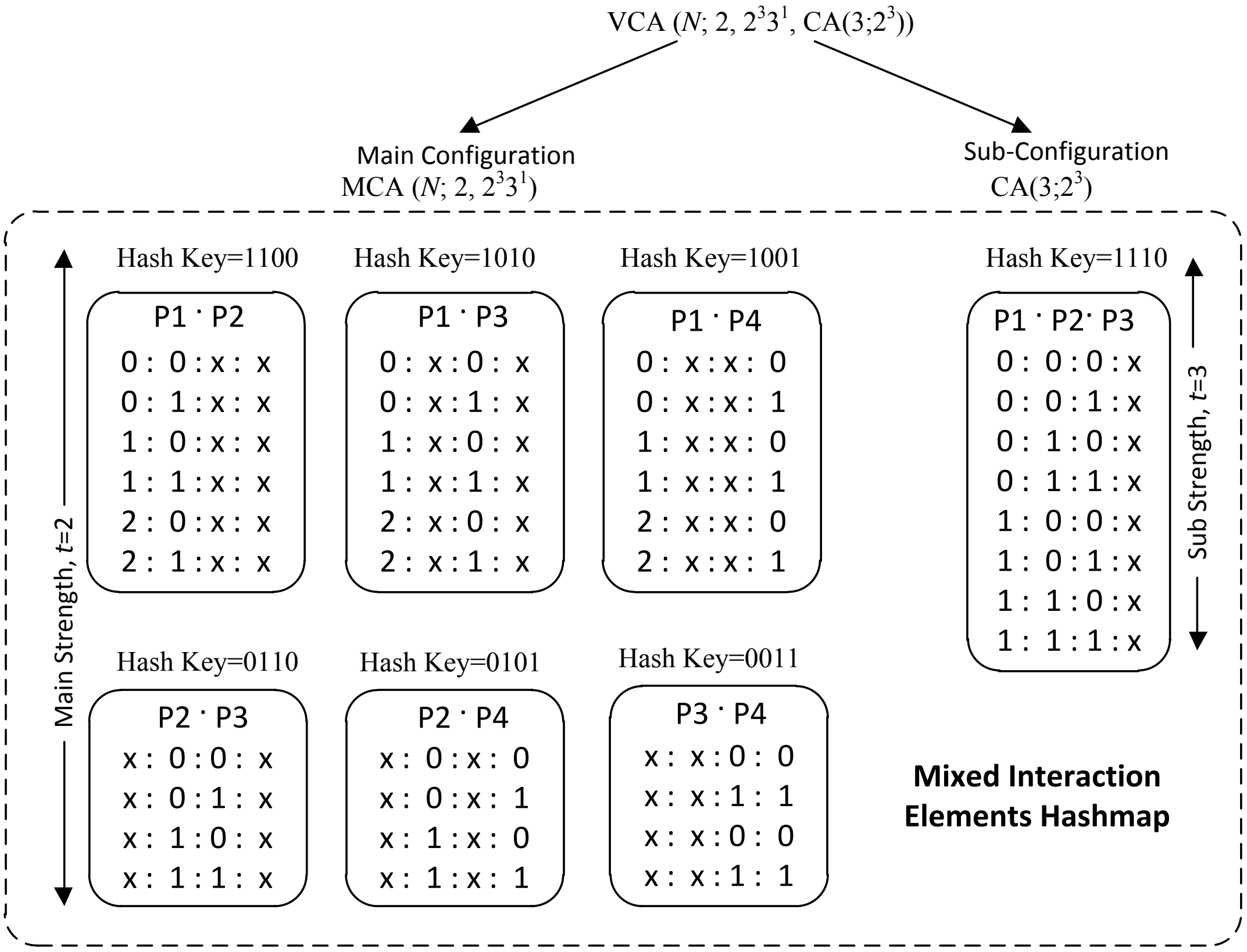

Figure 4. Hash Map and Interaction Elements for VCA ($N$; 2, $2^3$ $3^1$, *CA* (3; $2^3$))

Consider VCA ($N$; 2, $2^3$ $3^1$, CA (3; $2^3$)), as shown in Figure 4. The mixed- strength VCA comprises two parts: the main configuration MCA ($N$; 2, $2^3$ $3^1$) and sub-configuration CA (3; $2^3$), respectively. The main configuration, MCA ($N$; 2, $2^3$ $3^1$), requires a two-way interaction (as main strength) for a system of four parameters. The algorithm first generates all possibilities of binary numbers up to four digits. Subsequently, the binary numbers that contain two 1s are selected, indicating that a pairwise interaction (i.e., $t = 2$) exists. For example, the binary number 1100 refers to a $P_1$–$P_2$ interaction. $P_1$ has two values (0 and 1); $P_2$ has two values (0 and 1); $P_3$ has two values (0 and 1); and $P_4$ has three values (0, 1, and 2). The two-way parameter interaction has six possible combinations based on the parameter-generation algorithm. Approximately 2 × 3 possible interaction elements exist between $P_1$ and $P_4$ for the combination 1001, whereby $P_1$ and $P_4$ are available. For each parameter in the combination (i.e., with two 1s), the value of the corresponding parameter is included in the interaction elements. The excluded values are marked here as "don't care". This process is iteratively repeated for the other five interactions: ($P_1$, $P_2$), ($P_1$, $P_3$), ($P_2$, $P_3$), ($P_2$, $P_4$), and ($P_3$, $P_4$). The sub-configuration, CA (3; $2^3$), similarly requires a three-way interaction (as sub-strength) for a system of three parameters. A three-way interaction yields the ($P_1$, $P_2$, $P_3$) interaction.

The complete interaction elements of the overall VCA ($N$; 2, $2^3$ $3^1$, $CA$ (3; $2^3$)) are the combinations from both MCA ($N$; 2, $2^3$ $3^1$) and CA (3; $2^3$). The hash map list of mixed-interaction elements $H_s$, which employs the binary representation of the interaction as the map retrieval key, is implemented to ensure efficient indexing for storage and retrieval. The complete algorithm for the interaction element-generation algorithm is highlighted in Figure 5.

**Algorithm 3:** Algorithm for Interaction Elements Generation

**Input:** parameters' values $p$, strength of coverage $t$, sub strength of coverage $t_{sub}$
**Output:** $H_s$ representing the hash map list of $t - way\ interaction$ elements

```
Initialize H_s = ∅
Let m = max no. of defined parameters
Let p = {p_0, p_1, ..., p_j} where p is the set of values for each parameter
for index = 0 to 2^m−1 do
    Let b = binary number
    b = convert index to binary
    for i = 0 to length of b do
        if b[i] is equal to "1" then
            Add the representative interaction value for p[index]
        else
            Add don't care value for p[index]
    if (the no. of 1's in b = t) or (the no. of 1's in b = t_sub) then
        Set hashkey = b
        Put p into the hashmap H_s using the hashkey
Return H_s
```

Figure 5. Algorithm for Interaction Element-Generation

### 6.2 Test Suite Generation Algorithm based on ATLBO

ATLBO first initializes the population search space as a $D$-dimensional vector, $X_j = [X_{j,1}, X_{j,2}, X_{j,3} \ldots, X_{j,D}]$, where each dimension represents a parameter and contains integer numbers between 0 and $v_i$ (i.e., the value number of the *ith* parameter). TLBO requires both local and global search to be summoned per iteration. Conversely, ATLBO permits the adaptive selection of the local and global search through the fuzzy inference selection. The net effect is that ATLBO has less fitness function evaluation than the original TLBO for the same iteration number.

A discrete version of ATLBO is applied in the problem of generating $t$-way tests. Thus, each individual $X_j$ must capture the parameters as a valid range of integer numbers (i.e., based on user inputs) to deal with discrete parameters and values. Local and global search updates in ATLBO may result into the necessary rounding off of floating point values.

The rounding off of floating-point values should be addressed, as well as the out-of-range values. The *clamping rule* at the boundary within ATLBO is established to restrict parameter values for both lower and higher bounds. At least three possibilities exist in dealing with boundary conditions utilized in literature for discrete problems, i.e., invisible, reflecting, and absorbing walls [42]. When a current value exceeds the boundary in the invisible walls, the corresponding fitness value is not computed. When a value reaches the boundary in the reflecting walls, it is reflected back to the search space (i.e., mirroring effects). The boundary condition returns the current value to the search space by resetting its position to the other endpoint when the value moves out-of-range in the absorbing walls. For example, if a parameter value is in the range from 1 to 4, the position is reset to 1 when it reaches a value larger than 4.In this study, the absorbing-wall approach is used as the clamping rule in implementing ATLBO.

The local and global search processes of ATLBO are iteratively continued until convergence has been achieved (i.e., if and only if all the interaction elements from $H_s$ are completely removed), in relation to the stopping criteria. Two approaches are considered for storing and locating the interaction elements: array list and hash map. The array list approach is fast for small values but is not scalable for large parameters, because it must iterate the entire lists to fetch the required interaction values. Given that the process of fetching and locating the required interaction values are fundamentally important for fitness function evaluation, the array list approach can introduce time performance penalty. Alternatively, hash map offers an effective approach of locating the required interaction values utilizing only the unique key based on the binary interaction value itself. Thus, the hash map approach is considered for ATLBO. The ATLBO test suite generator is summarized in Figure 6 based on the aforementioned design choices.

**Algorithm 4:** The Adaptive TLBO for Mixed Strength t-way Test Suite Generation

**Input:** parameters' values $p$, strength of coverage $t$, and sub strength of coverage $t_{sub}$
**Output:** the final test suite $F_s$

Initialize random populations of learners $X$ and evaluate the mean of all learners $X_{mean}$
Initialize the required $t-way\ interaction$ elements in the hashmap $H_s$ based on the values of $p$, $t$, and $t_{sub}$
Define the member functions for the linguistic variables
Define the fuzzy rules
$X_{teacher}$ = Randomly select a learner from $X$
$X_{best} = X_{teacher}$
**while** *the hashmap $H_s$ is not empty* **do**
  **for** $i = 1$ *to population size* **do**
    Compute $Q_m$, $I_m$ and $D_m$
    Fuzzify based on $Q_m$, $I_m$ and $D_m$
    Defuzzify and set Selection = crisp output
    **if** *Selection* $> 50$ **then**
      /* Teacher Phase ... Exploration */
      Select $X_{teacher}$ and calculate $X_{mean}$
      $T_F = round(1 + r(0,1))$
      $X_i^{t+1} = X_i^t + r(X_{teacher} - T_F X_{mean})$
      **if** $f(X_i^{t+1})$ *is better than* $f(X_i^t)$ **then**
        $X_i^t = X_i^{t+1}$
    **else**
      /* Learner Phase ... Exploitation */
      Randomly select one learner $X_j^t$ from the population X such that $i \neq j$
      **if** $f(X_i^t)$ *is better than* $f(X_j^t)$ **then**
        $X_i^{t+1} = X_i^t + r(X_j^t - X_i^t)$
      **else**
        $X_i^{t+1} = X_i^t + r(X_i^t - X_j^t)$
      **if** $f(X_i^{t+1})$ *is better than* $f(X_i^t)$ **then**
        $X_i^t = X_i^{t+1}$
  Get best result ($X_{best}$) from the current population $X$ and put it in the final test suite list $F_s$
  Remove the interaction covered by $X_{best}$ from the hashmap $H_s$

Figure 6. ATLBO for Generating Mixed-Strength *t*-way Test Suite

## 7. Experiments

Our experiments focus on three related goals: (1) to characterize the generation efficiency and performance of ATLBO against the original TLBO (i.e., the efficiency is characterized by the size of the generated test suite, whereas the performance is characterized by the execution time of each strategy); (2) to gauge the adaptive distribution pattern of the exploration and exploitation for ATLBO; and (3) to benchmark ATLBO against other meta-heuristic approaches.

We divide our experiments into two parts to achieve the aforementioned goals. First , we adopt three selected CAs (i.e., CA(N; 2, $10^5$), CA(N; 2, $4^2$ $5^5$), and CA(N; 2, $2^3$ $3^5$)) and three selected VCAs (i.e., VCA(N; 2, $5^2$ $4^2$ $3^2$, CA (3, $4^2$ $3^2$)), VCA (N; 2, $5^7$, CA (3, $5^3$)), and VCA(N; 2, $3^{13}$, CA (3, $3^3$))) based on the interaction strength $t = 2$. With the first step, we highlight the time and size performance of the implemented ATLBO and original TLBO. Second, we benchmark the generated test suite sizes of the proposed ATLBO and TLBO implementation against each other and against existing meta-heuristic-based strategies based on the benchmark experiments published in [20]. Specifically, the benchmark experiments involve CA ($N$; $t$, $3^p$) with varying $t$ (2–4) and $p$ (2–12), CA($N$; $t$, $v^7$) along with CA($N$; $t$, $v^{10}$) with varying $t$ (2–4) and $v$ (4–6), VCA(N; 2, $3^{15}$, {C}), VCA(N; 3, $3^{15}$, {C}), and VCA(N; 2, $4^3$ $5^3$ $6^2$, {C}).

A fair comparison among each meta-heuristic-based strategy [43-46] is impossible because of the potentially different number of fitness function evaluations, variation in data structure, language implementation, and running environment. Furthermore, each meta-heuristic may require the specific control parameter settings (e.g., PSO-based strategies rely on inertia weight, as well as social and cognitive parameters, as parameters, whereas CS relies on its elitism probability). Given that the meta-heuristic-based strategy implementations are unavailable to us, we cannot modify the algorithm internal settings and can only run our own experiments in our running environment. We also implement the $t$-way strategy based on the original TLBO for comparative purposes in the context of our study. A direct comparative performance of ATLBO with the original TLBO (i.e., even with the same iteration number) is also unfair. The original TLBO has twice as much fitness function evaluations compared with ATLBO with the same iteration number because of the serial execution of both exploration and exploitation steps. Thus, the iteration number within TLBO must always be half of ATLBO for a fair comparison.

We set the population size to 40 and the maximum iteration to 100 for ATLBO in all our experiments. We adopt the same population size for the original TLBO but with a maximum iteration of 50. Our ATLBO and TLBO implementations are based on the Java programming language. The experimental platform we employ comprises of a PC that runs on Windows 10, CPU 2.9 GHz Intel Core i5, 16 GB 1867 MHz DDR3 RAM, and a 512 MB flash hard-disk drive. We execute ATLBO and TLBO 30 times in all the experiments to ensure statistical significance. The best and mean times (whenever applicable), as well as the best and mean test sizes for each experiment are reported together. The best cell entries are marked as "*", whereas the best mean cell entries are marked in bold font. Cell entries that are unavailable are marked with a dash "-". We also track the mean percentage of exploration (i.e., global search) and exploration (i.e., local search) for each experiment that involves ATLBO to highlight the actual search progresses for different CAs and VCAs.

### 7.1 Characterizing Time and Size Performances for TLBO and ATLBO

Given that both implementations are based on the same data structure, language implementation, running environment, and fitness function evaluation, we can fairly compare the sizes and time performance for TLBO and ATLBO. Table 3 highlights our results, whereas Figure 7 depicts the mean exploration and exploitation percentage for ATLBO based on the provided CAs and VCAs.

Table 3. Characterizing TLBO and ATLBO

| ID | CA and VCA | Original TLBO | | | | ATLBO | | | | | |
|---|---|---|---|---|---|---|---|---|---|---|---|
| | | Size | | Time (s) | | Size | | Time (s) | | % Mean Exploit | %Mean Explore |
| | | Best | Mean | Best | Mean | Best | Mean | Best | Mean | | |
| CA1 | CA (N; 2, $10^5$) | 117 | 118.7 | 28.80 | 41.10 | 116* | **118.53** | 23.76* | **28.13** | 79.81 | 20.19 |
| CA2 | CA (N; 2, $4^2$ $5^5$) | 32 | 34.00 | 9.19* | **10.20** | 28* | **28.95** | 11.55 | 13.83 | 62.18 | 37.82 |
| CA3 | CA (N; 2, $2^3$ $3^5$) | 13 | 14.77 | 5.12* | **6.15** | 13 | **14.16** | 6.64 | 8.07 | 32.20 | 67.80 |
| VCA1 | VCA (N; 2, $5^2$ $4^2$ $3^2$, CA (3, $4^2$ $3^2$)) | 104 | **107.67** | 40.87* | **47.36** | 103* | 107.90 | 74.18 | 66.11 | 13.20 | 86.80 |
| VCA2 | VCA (N; 2, $5^7$, CA (3, $5^3$)) | 125 | 125.00 | 66.63* | **69.10** | 125 | **125.00** | 125.02 | 131.42 | 18.69 | 81.31 |
| VCA3 | VCA (N; 2, $3^{13}$, CA (3, $3^3$)) | 27 | 27.26 | 45.94* | **49.60** | 27 | **27.23** | 64.37 | 69.98 | 23.43 | 76.57 |

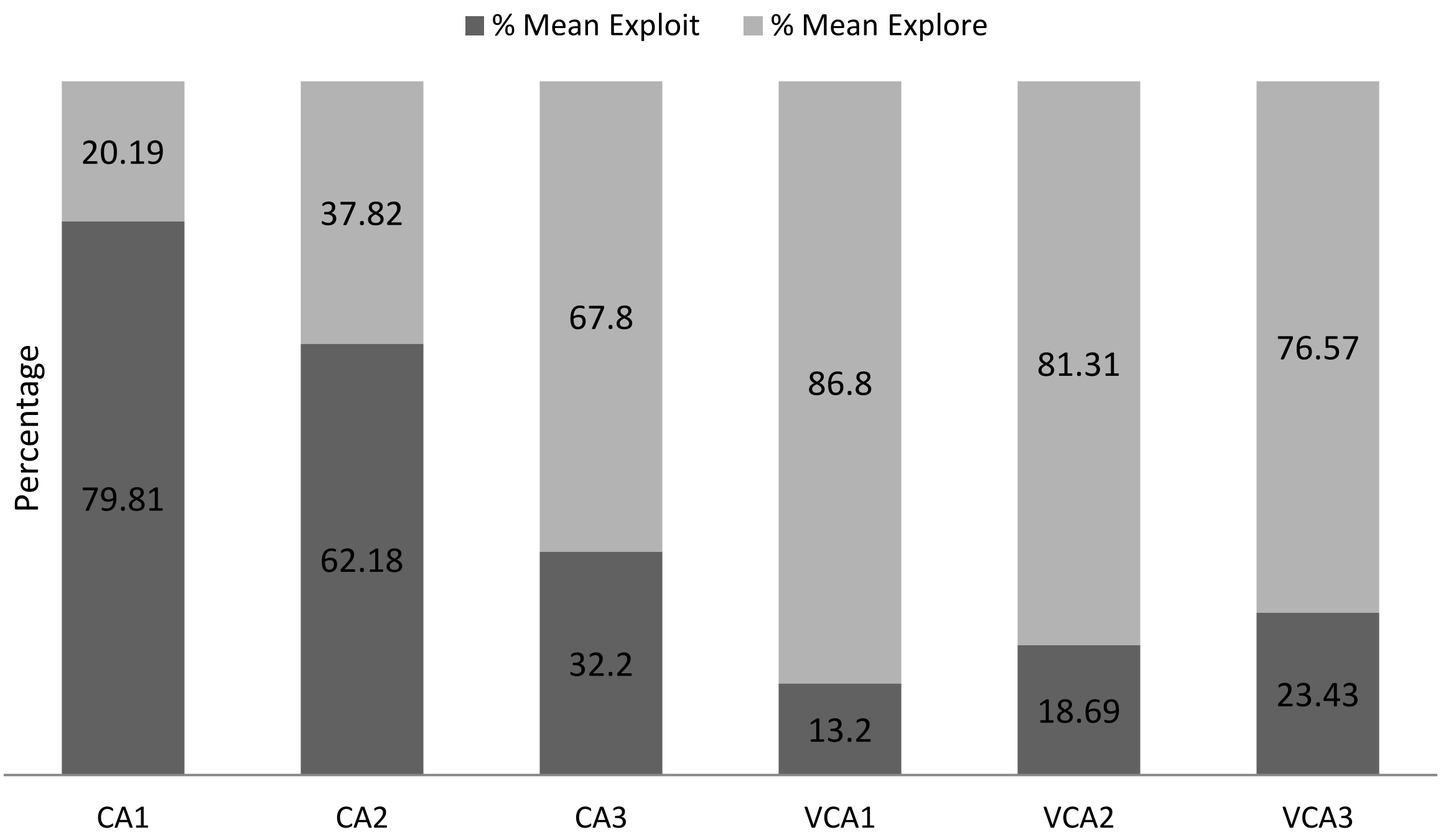

Figure 7. Mean Exploration and Exploitation Percentages of ATLBO for Table 3

### 7.2 Benchmarking with other Meta-Heuristic Strategies

Unlike the experiments in the previous section, the benchmark experiments in this section also include the ATBLO performance against all other strategies. However, the execution time is omitted because of the differences in the parameter control settings (e.g., maximum iteration and unequal evaluation of fitness function) and implementation (e.g., data structure and implementation language). Despite these differences, we believe that our comparison is still valid because the published best and mean test sizes are obtained utilizing the best control parameter settings.

Tables 4 to 9 highlight our results, whereas Figures 8 to 13 depict the mean exploration and exploitation percentage for ATLBO based on the provided CAs and VCAs.

Table 4. CA ($N$; $t$, $3^p$)

| $t$ | $p$ | PSTG [47] | | DPSO [20] | | APSO [12] | | CS [19] | | Original TLBO | | ATLBO | | | |
|---|---|---|---|---|---|---|---|---|---|---|---|---|---|---|---|
| | | Best | Mean | Best | Mean | Best | Mean | Best | Mean | Best | Mean | Best | Mean | % Mean Exploit | % Mean Explore |
| 2 | 4 | 9* | 10.15 | 9* | **9.00** | 9* | 9.95 | 9* | 10.0 | 9 | **9.00** | 9* | **9.00** | 96.36 | 3.64 |
| | 5 | 12 | 13.81 | 11* | 11.53 | 11* | 12.23 | 11* | 11.80 | 11* | 11.43 | 11* | **11.33** | 55.14 | 44.86 |
| | 6 | 13 | 15.11 | 14 | 14.50 | 12* | **13.78** | 13 | 14.20 | 13 | 14.60 | 13 | 14.33 | 53.49 | 46.51 |
| | 7 | 15* | 16.94 | 15* | 15.17 | 15* | 16.62 | 14* | 15.60 | 15* | 15.07 | 15* | **15.05** | 52.71 | 47.29 |
| | 8 | 15* | 17.57 | 15* | 16.00 | 15* | 16.92 | 15* | 15.80 | 15* | **15.70** | 15* | 15.90 | 40.88 | 59.12 |
| | 9 | 17 | 19.38 | 15* | 16.43 | 16 | 18.31 | 16 | 17.20 | 15* | 16.23 | 15* | **15.03** | 41.46 | 58.54 |
| | 10 | 17 | 19.78 | 16* | **17.30** | 17 | 18.12 | 17 | 17.80 | 16* | 17.40 | 16* | 17.37 | 37.02 | 62.98 |
| | 11 | 17 | 20.16 | 17 | 17.70 | - | - | 18 | 18.60 | 16* | 17.73 | 16* | **17.67** | 36.77 | 63.23 |
| | 12 | 18 | 21.34 | 16* | 17.93 | - | - | 18 | 18.80 | 17 | 18.10 | 17 | **17.80** | 37.14 | 62.86 |
| 3 | 5 | 39 | 41.37 | 41 | 43.17 | 41 | 42.20 | 38* | 39.20 | 38* | 42.53 | 38* | **42.37** | 61.59 | 38.41 |
| | 6 | 45 | 46.76 | 33* | **38.30** | 45 | 46.51 | 43 | 44.20 | 33* | 38.87 | 33* | 38.43 | 55.86 | 44.14 |
| | 7 | 50 | 52.20 | 48* | 50.43 | 48* | 51.12 | 48* | 50.40 | 50 | 50.53 | 49 | **50.37** | 40.27 | 59.73 |
| | 8 | 54 | 56.76 | 52 | 53.83 | 50* | 54.86 | 53 | 54.80 | 52 | **53.17** | 52 | 53.33 | 38.39 | 61.61 |
| | 9 | 58 | 60.30 | 56 | 57.77 | 59 | 60.21 | 58 | 59.80 | 56 | 57.77 | 55* | **57.50** | 35.01 | 64.99 |
| | 10 | 62 | 63.95 | 59* | 60.87 | 63 | 64.33 | 62 | 63.60 | 60 | 60.93 | 59* | **60.73** | 34.09 | 65.91 |
| | 11 | 64 | 65.68 | 63 | 63.97 | - | - | 66 | 68.20 | 62* | 63.70 | 62* | **63.57** | 32.17 | 67.83 |
| | 12 | 67 | 68.23 | 65* | 66.83 | - | - | 70 | 71.80 | 65* | 66.70 | 65* | **66.53** | 29.93 | 70.07 |
| 4 | 6 | 133 | 135.31 | 131 | 134.37 | 129* | 133.98 | 132 | 134.20 | 130 | **133.63** | 130 | 134.10 | 50.50 | 49.50 |
| | 7 | 155 | 158.12 | 150 | **155.23** | 154 | 157.42 | 154 | 156.80 | 146* | 155.77 | 152 | 156.03 | 40.22 | 59.78 |
| | 8 | 175 | 176.94 | 171* | 175.60 | 178 | 179.70 | 173 | 174.80 | 171* | 175.83 | 171* | **175.50** | 33.85 | 66.15 |
| | 9 | 195 | 198.72 | 187 | 192.27 | 190 | 194.13 | 195 | 197.80 | 187 | 190.33 | 156* | **189.60** | 31.76 | 68.24 |
| | 10 | 210 | 212.71 | 206 | 219.07 | 214 | 212.21 | 211 | 212.20 | 205* | 208.80 | 207 | **208.43** | 27.20 | 72.80 |
| | 11 | 222 | 226.59 | 221 | 224.27 | - | - | 229 | 231.00 | 221* | **224.12** | 221* | 223.43 | 24.65 | 75.35 |
| | 12 | 244 | 248.97 | 237 | 239.83 | - | - | 253 | 255.80 | 236 | 239.29 | 235* | **237.83** | 22.41 | 77.59 |

Table 5. CA ($N$; $t$, $v^7$)

| $t$ | $v$ | PSTG [47] | | DPSO [20] | | APSO [12] | | CS [19] | | Original TLBO | | ATLBO | | | |
|---|---|---|---|---|---|---|---|---|---|---|---|---|---|---|---|
| | | Best | Mean | Best | Mean | Best | Mean | Best | Mean | Best | Mean | Best | Mean | % Mean Exploit | %Mean Explore |
| *2* | 2 | 6* | 6.82 | 7 | 7.00 | 6* | **6.73** | 6* | 6.80 | 7 | 7.00 | 7 | 7.00 | 50.87 | 49.13 |
| | 3 | 15 | 15.23 | 14* | **15.00** | 15 | 15.56 | 15 | 16.20 | 15 | 15.10 | 15 | 15.07 | 51.60 | 48.40 |
| | 4 | 26 | 27.22 | 24 | 25.33 | 25 | 26.36 | 25 | 26.40 | 24 | 25.27 | 23* | **25.17** | 57.74 | 42.26 |
| | 5 | 37 | 38.14 | 34* | 35.47 | 35 | 37.92 | 37 | 38.60 | 34* | **35.43** | 34* | 35.47 | 63.82 | 36.18 |
| *3* | 2 | 13 | **13.61** | 15 | 15.06 | 15 | 15.80 | 12* | 13.80 | 15 | 15.12 | 15 | 15.12 | 48.42 | 51.58 |
| | 3 | 50 | 51.75 | 49 | 50.60 | 48* | 51.12 | 49 | 51.60 | 49 | 50.38 | 49 | **50.29** | 39.60 | 60.40 |
| | 4 | 116 | 118.13 | 112 | **115.27** | 118 | 120.41 | 117 | 118.40 | 112 | 115.37 | 111* | 115.67 | 43.16 | 56.84 |
| | 5 | 225 | 227.21 | 216* | **219.20** | 239 | 243.29 | 223 | 225.40 | 217* | 219.90 | 216* | 219.40 | 44.77 | 55.23 |
| *4* | 2 | 29 | 31.49 | 34 | 34.00 | 30 | 31.34 | 27* | **29.60** | 31 | 33.70 | 31 | 33.68 | 46.04 | 53.96 |
| | 3 | 155 | 157.77 | 150* | **154.73** | 153 | 155.20 | 155 | 156.80 | 151 | 155.25 | 150* | 155.24 | 39.42 | 60.58 |
| | 4 | 487 | 489.91 | 472* | 481.53 | 472 | **478.90** | 487 | 490.20 | 480 | 485.53 | 478 | 484.69 | 39.90 | 60.10 |
| | 5 | 1176 | 1180.63 | 1148* | **1155.63** | 1162 | 1169.94 | 1171 | 1175.20 | 1166 | 1173.17 | 1166 | 1173.45 | 40.14 | 59.86 |

Table 6. CA ($N$; $t$, $v^{10}$)

| $t$ | $v$ | PSTG [47] | | DPSO [20] | | CS [19] | | Original TLBO | | ATLBO | | | |
|---|---|---|---|---|---|---|---|---|---|---|---|---|---|
| | | Best | Mean | Best | Mean | Best | Mean | Best | Mean | Best | Mean | %Mean Exploit | %Mean Explore |
| 2 | 4 | - | - | 28* | 29.20 | - | - | 28* | **28.73** | 28* | 28.69 | 42.42 | 57.58 |
| | 5 | 45 | 48.31 | 42 | 43.67 | 45 | 47.8 | 41* | **43.30** | 42 | 43.53 | 46.92 | 53.08 |
| | 6 | - | - | 58* | **59.23** | - | - | 58* | 59.47 | 58* | 59.33 | 50.27 | 49.73 |
| 3 | 4 | - | - | 141 | 143.70 | - | - | 140* | 142.57 | 140* | 142.80 | 30.77 | 69.23 |
| | 5 | 287 | 298.00 | 273 | 276.20 | 297 | 299.20 | 273 | 275.70 | 272* | **275.23** | 31.04 | 68.96 |
| | 6 | - | - | 467 | 470.50 | - | - | 467 | 470.47 | 466* | **469.90** | 31.53 | 68.47 |
| 4 | 4 | - | - | 664 | 667.00 | - | - | 663 | 668.12 | 661* | **664.06** | 25.68 | 74.32 |
| | 5 | 1716 | 1726.72 | 1618* | **1620.80** | 1731 | 1740.20 | 1621 | 1621.80 | 1619 | 1620.91 | 22.32 | 77.68 |
| | 6 | - | - | 3339 | 3342.50 | - | - | 3338* | 3343.81 | 3338* | **3342.10** | 21.13 | 78.87 |

Table 7. VCA (N; 2, $3^{15}$, {C})

| ID | VCA | PSTG [47] | | DPSO [20] | | ACS [16] | | SA [17] | | Original TLBO | | ATLBO | | | |
|---|---|---|---|---|---|---|---|---|---|---|---|---|---|---|---|
| | | Best | Mean | Best | Mean | Best | Mean | Best | Mean | Best | Mean | Best | Mean | %Mean Exploit | %Mean Explore |
| VCA1 | Ø | 19 | 20.92 | 18 | **18.63** | 19 | - | 16* | - | 19 | 19.67 | 18 | 19.30 | 31.30 | 68.70 |
| VCA2 | CA (3, $3^3$) | 27* | 27.50 | 27* | 27.27 | 27* | - | 27* | - | 27* | 27.33 | 27* | **27.00** | 22.26 | 77.74 |
| VCA3 | CA (3, $3^3$)$^2$ | 27* | 27.94 | 27* | 27.83 | 27* | - | 27* | - | 27* | **27.47** | 27* | 27.53 | 21.46 | 78.54 |
| VCA4 | CA (3, $3^3$)$^3$ | 27* | 28.13 | 27* | 28.00 | 27* | - | 27* | - | 27* | 27.93 | 27* | **27.43** | 22.00 | 78.00 |
| VCA5 | CA (3, $3^4$) | 30 | 31.47 | 27* | 31.43 | 27* | - | 27* | - | 27* | 32.73 | 27* | **27.00** | 22.26 | 77.74 |
| VCA6 | CA (3, $3^5$) | 38 | **39.83** | 38 | 40.93 | 38 | - | 33* | - | 38 | 40.97 | 38 | 40.60 | 16.25 | 83.75 |
| VCA7 | CA (3, $3^6$) | 45 | 46.42 | 43 | 45.70 | 45 | - | 34* | - | 43 | 43.73 | 43 | **43.67** | 18.05 | 81.95 |
| VCA8 | CA (3, $3^7$) | 49 | 51.68 | 47* | 49.87 | 48 | - | 41 | - | 49 | 50.03 | 47* | **49.83** | 17.96 | 82.04 |
| VCA9 | CA (4, $3^4$) | 81* | 82.21 | 81* | **81.03** | - | - | - | - | 81* | **81.03** | 81* | **81.03** | 7.44 | 92.56 |
| VCA10 | CA (4, $3^5$) | 97 | 99.31 | 85* | **94.50** | - | - | - | - | 89 | 97.53 | 87 | 96.90 | 7.26 | 92.74 |
| VCA11 | CA (4, $3^7$) | 158 | 160.31 | 152* | 156.83 | - | - | - | - | 153 | 156.51 | 152* | **156.33** | 10.74 | 89.26 |

Table 8. VCA (N; 3, $3^{15}$, {C})

| ID | VCA | PSTG [47] | | DPSO [20] | | HSS [18] | | Original TLBO | | ATLBO | | | |
|---|---|---|---|---|---|---|---|---|---|---|---|---|---|
| | | Best | Mean | Best | Mean | Best | Mean | Best | Mean | Best | Mean | %Mean Exploit | %Mean Explore |
| VCA1 | Ø | 75 | 78.69 | 72* | **73.97** | 75 | 75.00 | 73 | 74.47 | 73 | 74.37 | 24.64 | 75.36 |
| VCA2 | CA (4, $3^4$) | 91 | 91.80 | 86 | 89.83 | 87 | 87.00 | 90 | 90.03 | 85* | **89.23** | 20.36 | 79.64 |
| VCA3 | CA (4, $3^4$)$^2$ | 91 | 92.21 | 88 | 90.77 | 90 | 90.00 | 86* | **89.76** | 87 | 90.10 | 20.24 | 79.76 |
| VCA4 | CA (4, $3^5$) | 114 | 117.30 | 107 | **111.17** | 112 | 112.00 | 106* | 111.90 | 107 | 112.13 | 16.44 | 83.56 |
| VCA5 | CA (4, $3^7$) | 159 | 162.23 | 152* | 158.57 | 159 | 160.10 | 155 | 158.40 | 153 | **158.30** | 12.11 | 87.89 |
| VCA6 | CA (4, $3^9$) | 195 | 199.28 | 193 | 196.00 | 199 | 199.80 | 190 | 193.40 | 189* | **193.29** | 11.15 | 88.85 |
| VCA7 | CA (4, $3^{11}$) | 226 | 230.64 | 225* | 227.50 | 242 | 243.00 | 226 | 229.51 | 225* | **227.48** | 10.01 | 89.99 |

Table 9. VCA (N; 2, $4^3$ $5^3$ $6^2$, {C})

| ID | VCA | PSTG [47] | | DPSO [20] | | HSS [18] | | ACS [16] | | SA [17] | | Original TLBO | | ATLBO | | | |
|---|---|---|---|---|---|---|---|---|---|---|---|---|---|---|---|---|---|
| | | Best | Mean | Best | Mean | Best | Mean | Best | Mean | Best | Mean | Best | Mean | Best | Mean | %Mean Exploit | %Mean Explore |
| VCA1 | Ø | 42 | 43.60 | 40 | 42.30 | 42 | 43.50 | 41 | - | 36* | - | 40 | 42.03 | 39 | 41.63 | 43.47 | 56.53 |
| VCA2 | CA (3, $4^3$) | 64* | 65.50 | 64* | **64.00** | 64* | **64.00** | 64 | - | 64 | - | 64* | 64.03 | 64* | 64.03 | 31.11 | 68.89 |
| VCA3 | CA (3, $4^3$ $5^2$) | 124 | 126.60 | 119 | 124.70 | 116 | **120.90** | 104 | - | 100* | - | 121 | 125.67 | 122 | 124.5 | 18.31 | 81.69 |
| VCA4 | CA (3, $4^3$), CA (3, $5^3$) | 125* | 127.90 | 125* | **125.00** | 125* | **125.00** | 125* | - | 125* | - | 125* | **125.00** | 125* | **125.00** | 15.88 | 84.12 |
| VCA5 | CA (3, $4^3$ $5^3$ $6^1$) | 206 | 210.20 | 203 | 207.50 | 212 | 214 | 201 | - | 171* | - | 203 | 208.77 | 203 | **208.68** | 14.42 | 85.58 |
| VCA6 | CA (3, $4^3$), CA (4,$5^3$ $6^1$) | 750 | 755.70 | 750* | 750.80 | 750* | **750.00** | - | - | - | - | 750* | **750.00** | 750* | **750.00** | 12.70 | 87.30 |
| VCA7 | CA (4, $4^3$ $5^2$) | 472 | 478.10 | 440* | **450.60** | 453 | 454.3 | - | - | - | - | 459 | 466.70 | 451 | 459.10 | 6.52 | 93.48 |

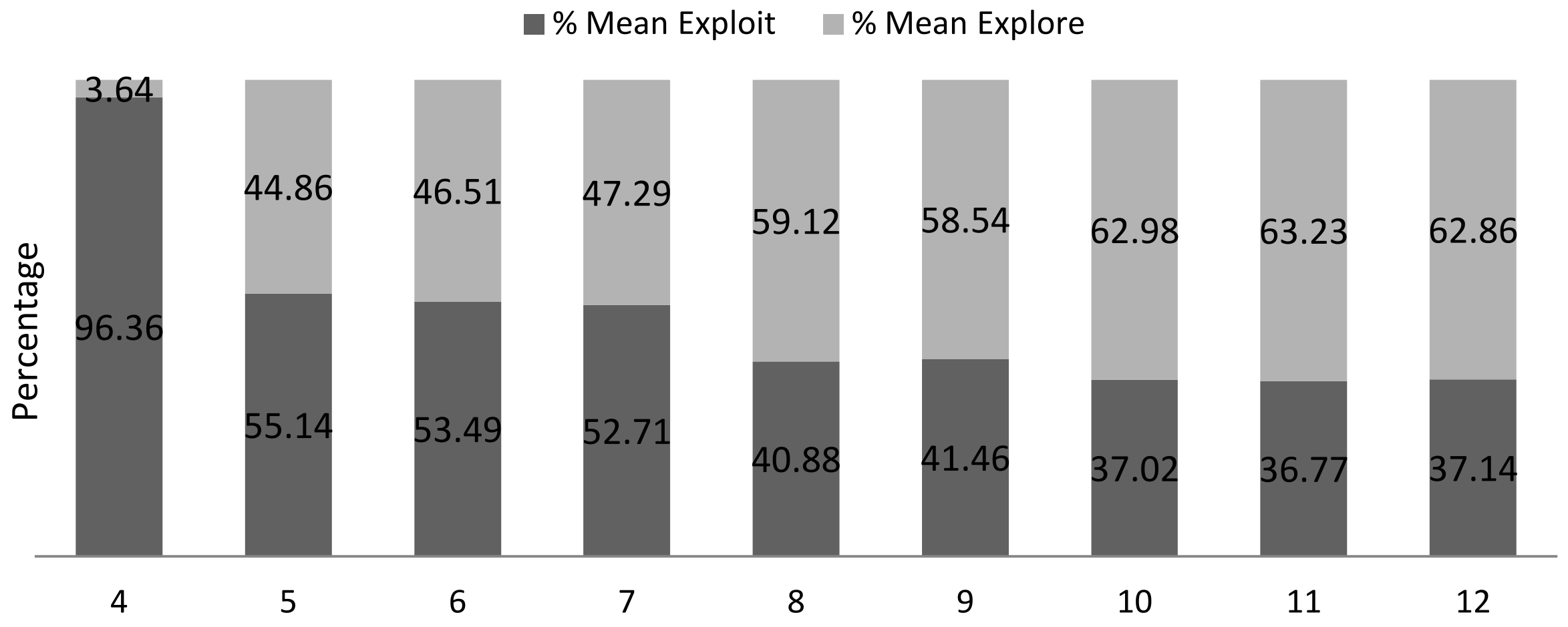

a) CA ($N$; $t$, $3^p$) with $t$ = 2, $p$ varies from 4 to 12

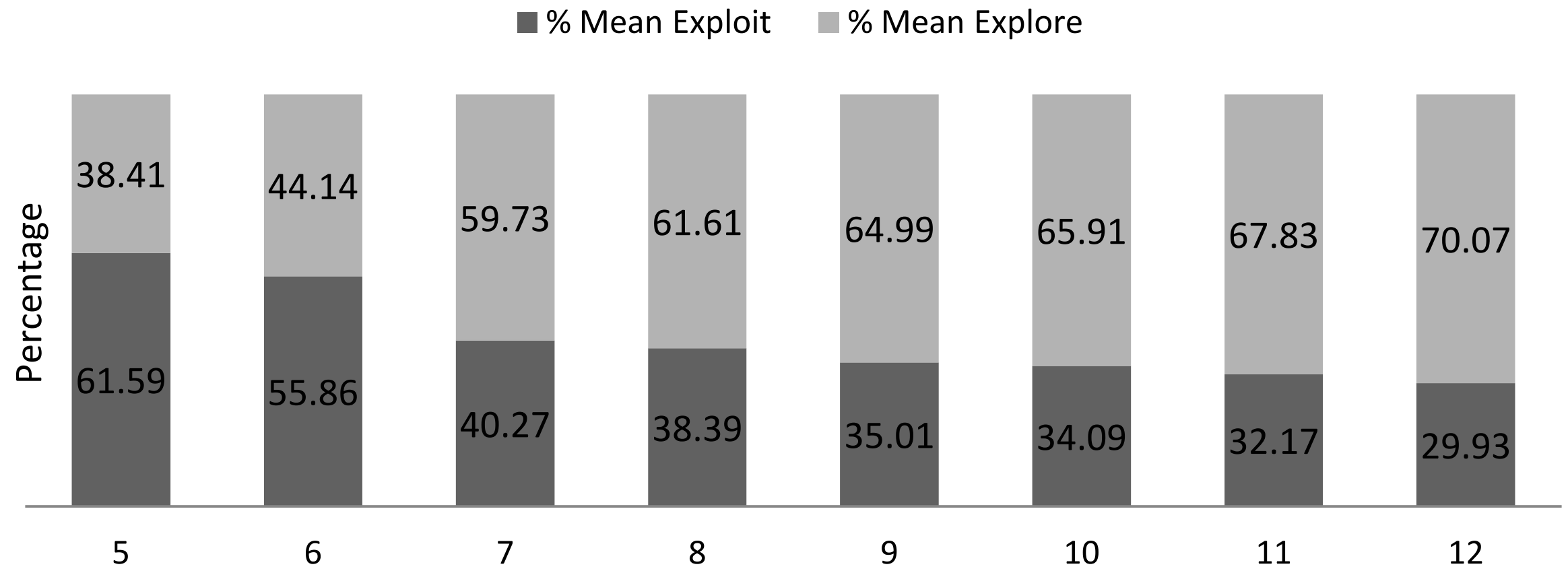

b) CA ($N$; $t$, $3^p$) with $t$ = 3, $p$ varies from 5 to 12

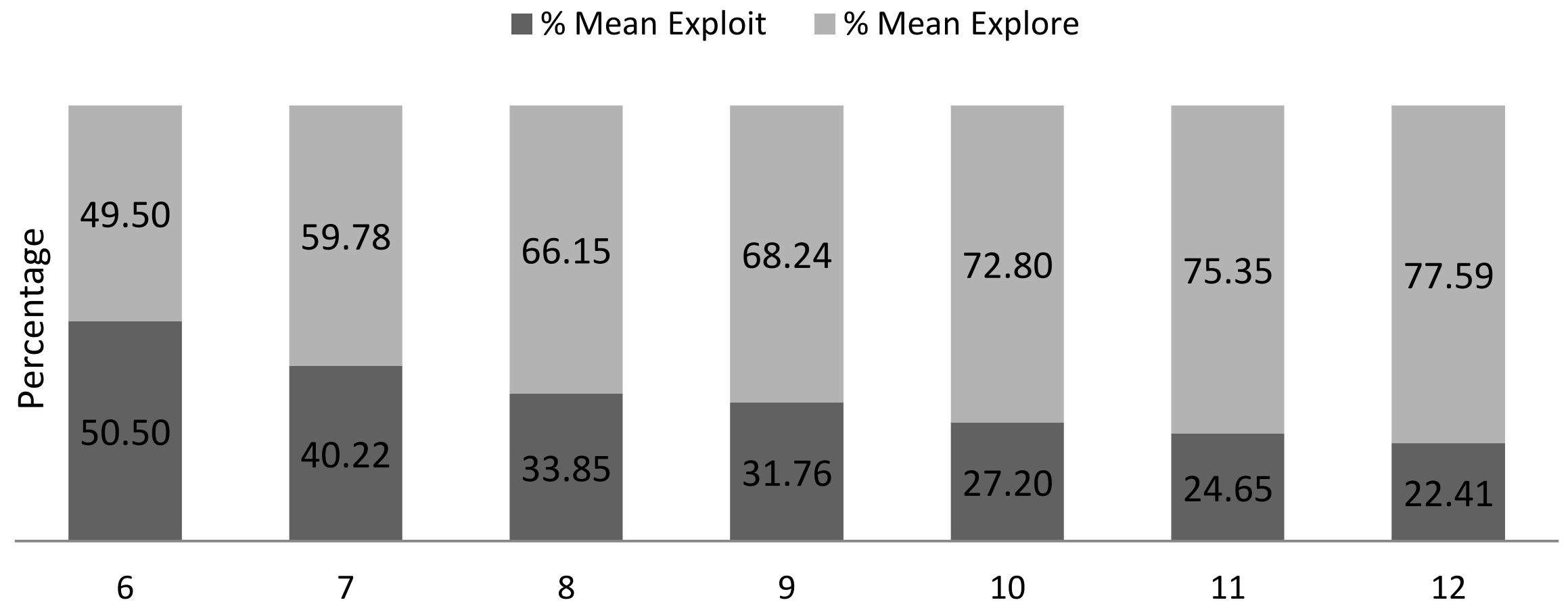

c) CA ($N$; $t$, $3^p$) with $t$ = 4, $p$ varies from 6 to 12

Figure 8. Mean Exploration and Exploitation Percentages of ATLBOfor Table 4

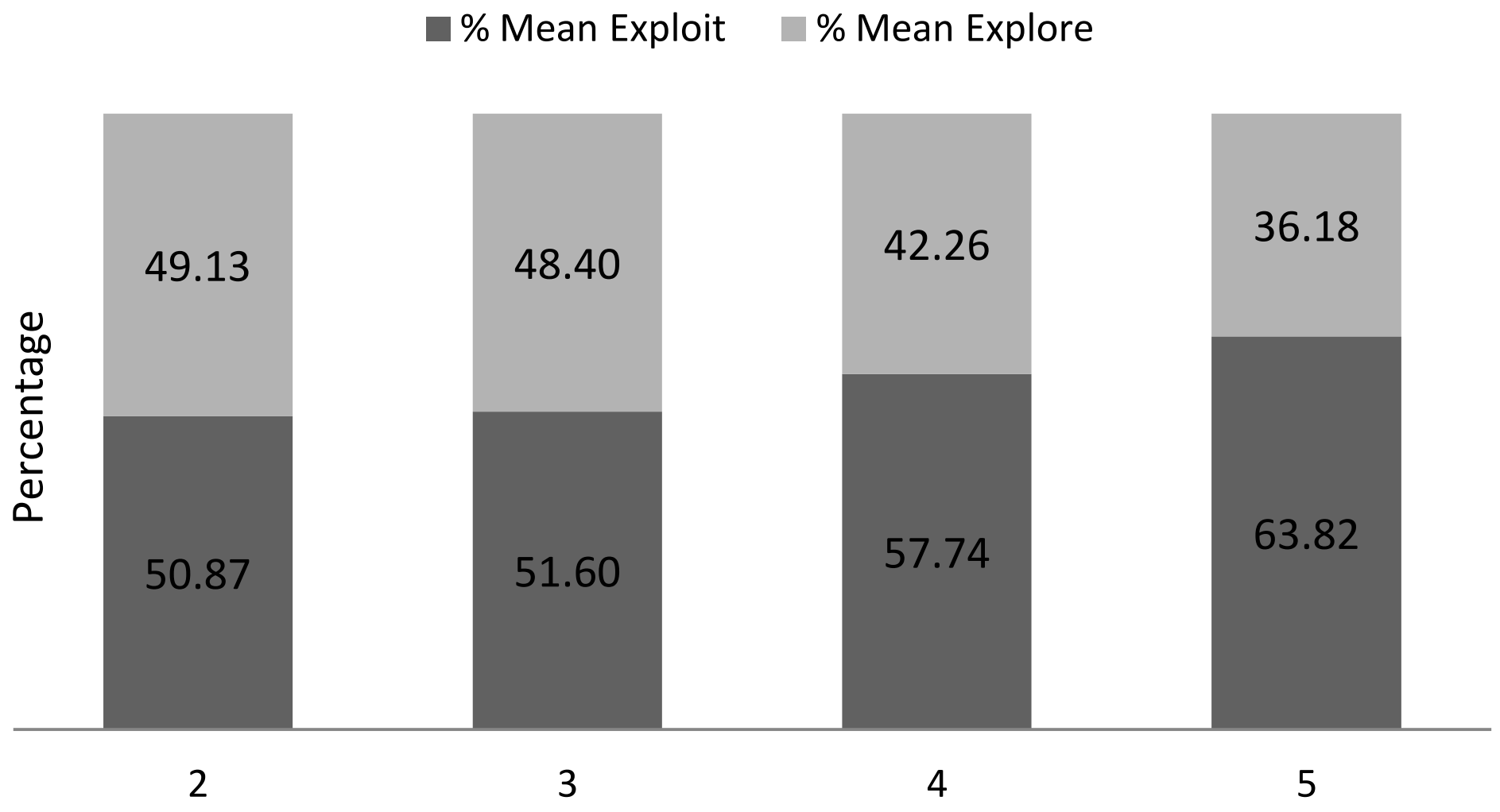

a) CA ($N$; $t$, $v^7$) with $t$ = 2, $v$ varies from 2 to 5

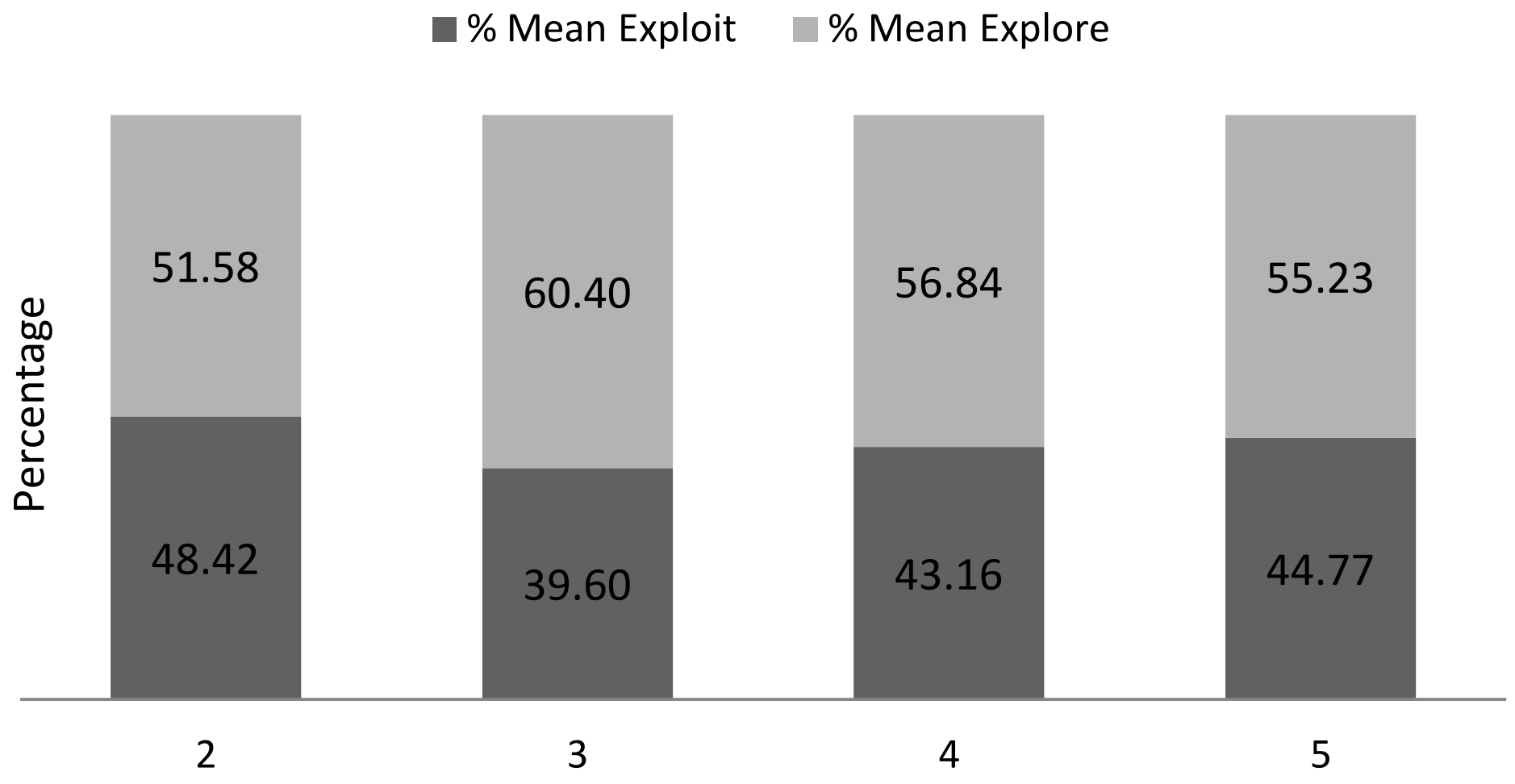

b) CA ($N$; $t$, $v^7$) with $t$ = 3, $v$ varies from 2 to 5

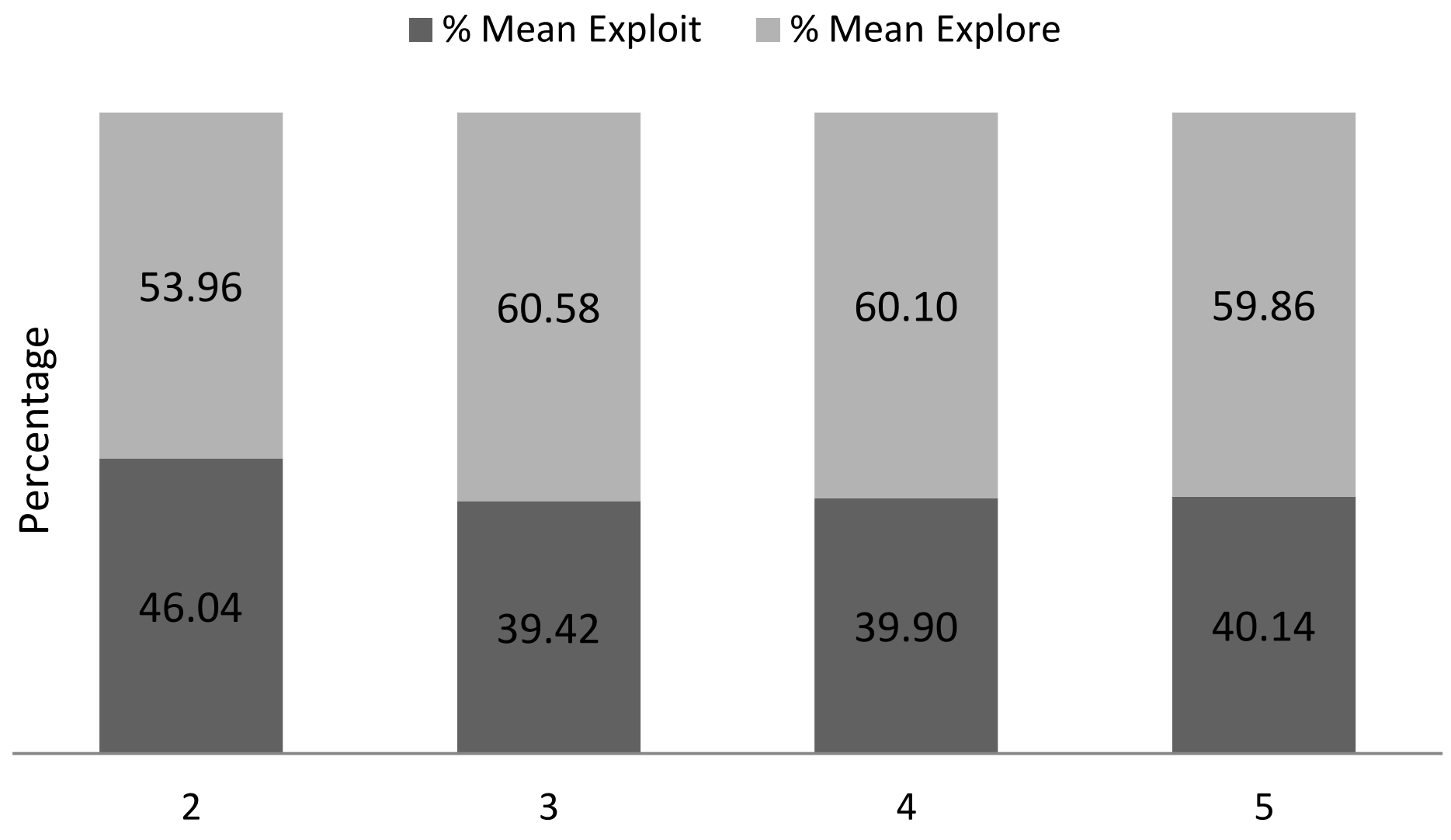

c) CA ($N$; $t$, $v^7$) with $t$ = 4, $v$ varies from 2 to 5

Figure 9. Mean Exploration and Exploitation Percentages of ATLBO for Table 5

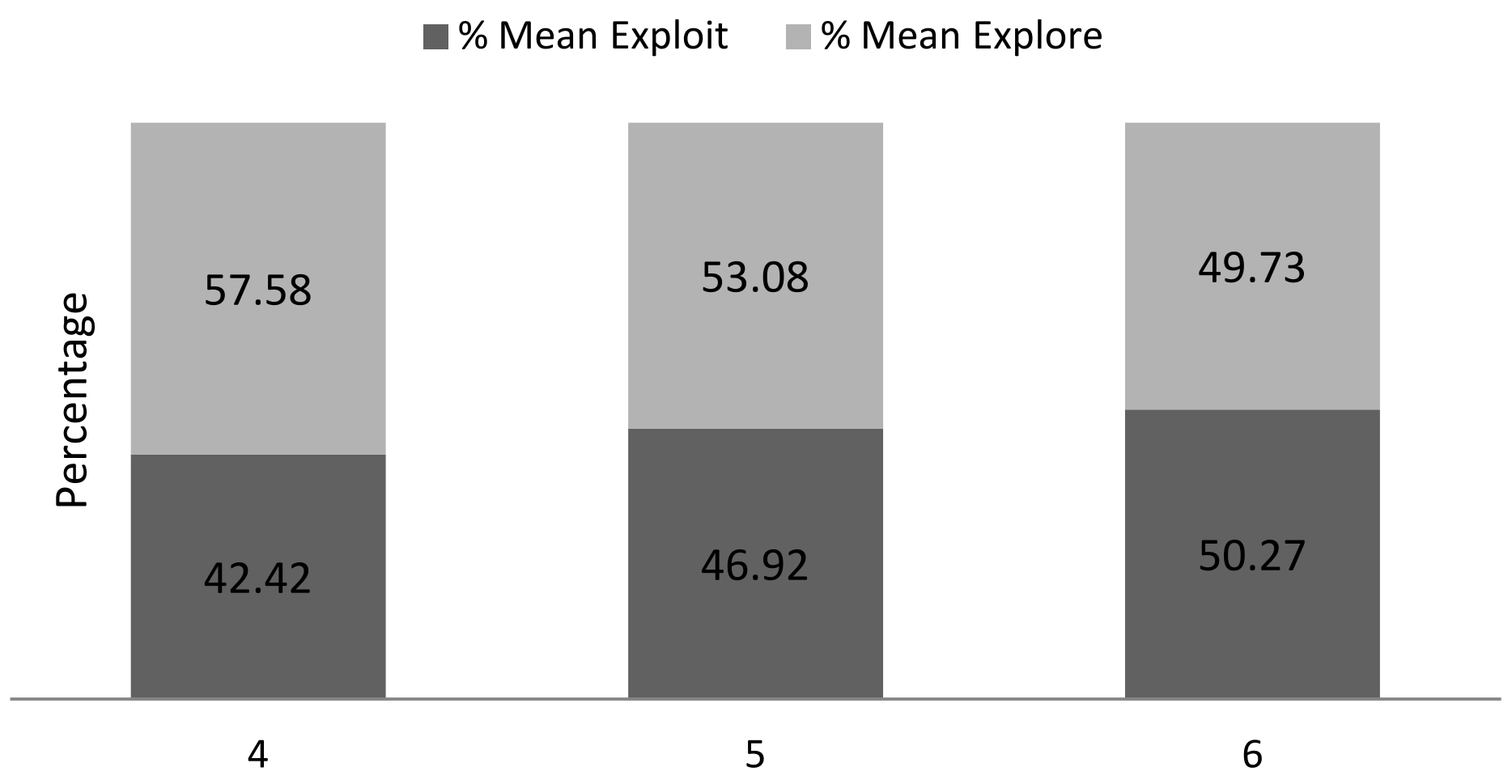

a) CA ($N$; $t$, $v^{10}$) with $t$ = 2, $v$ varies from 4 to 6

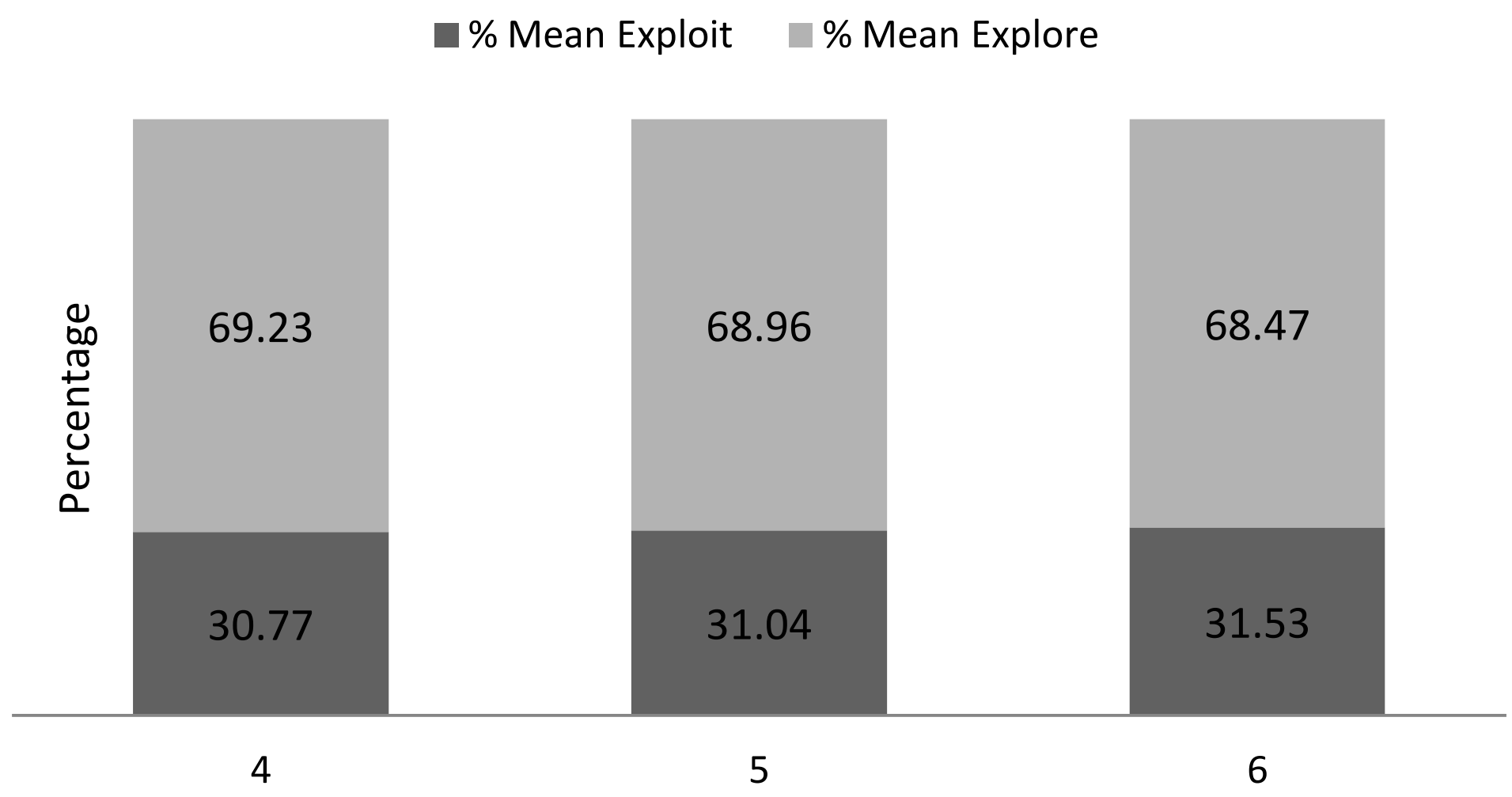

b) CA ($N$; $t$, $v^{10}$) with $t$ = 3, $v$ varies from 4 to 6

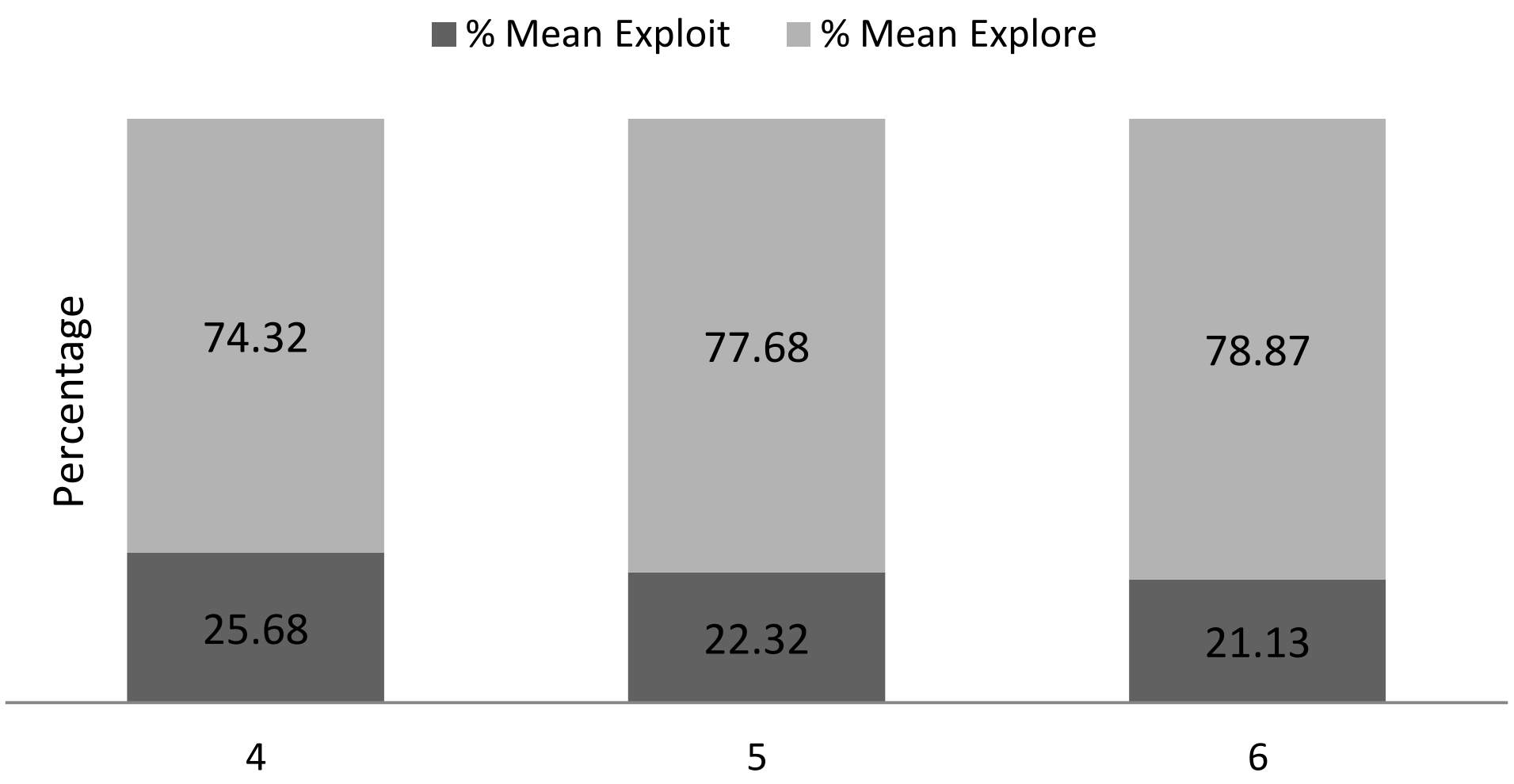

c) CA ($N$; $t$, $v^{10}$) with $t$ = 4, $v$ varies from 4 to 6

Figure 10. Mean Exploration and Exploitation Percentages of ATLBO for Table 6

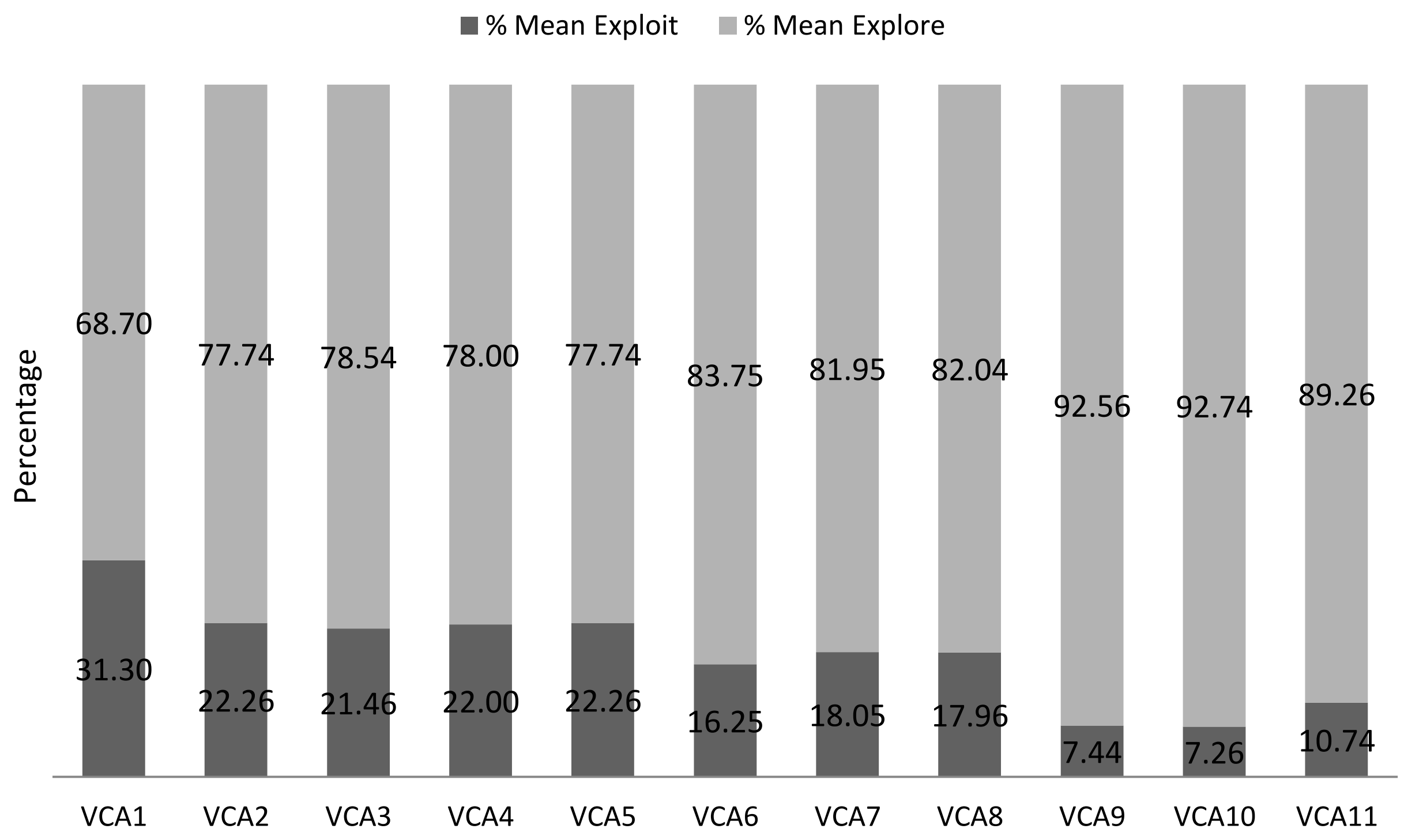

Figure 11. Mean Exploration and Exploitation Percentages of ATLBO for Table 7

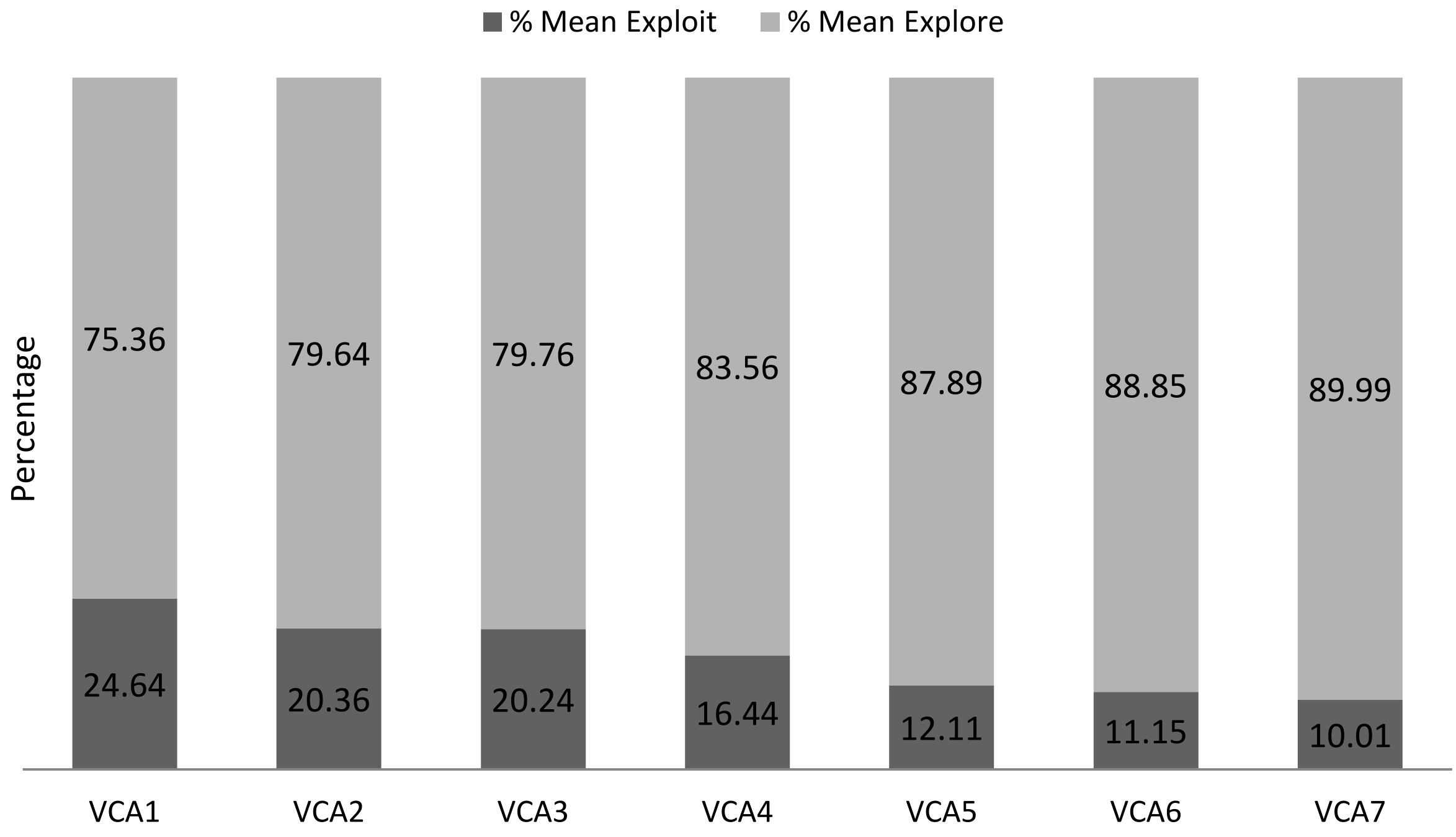

Figure 12. Mean Exploration and Exploitation Percentages of ATLBO for Table 8

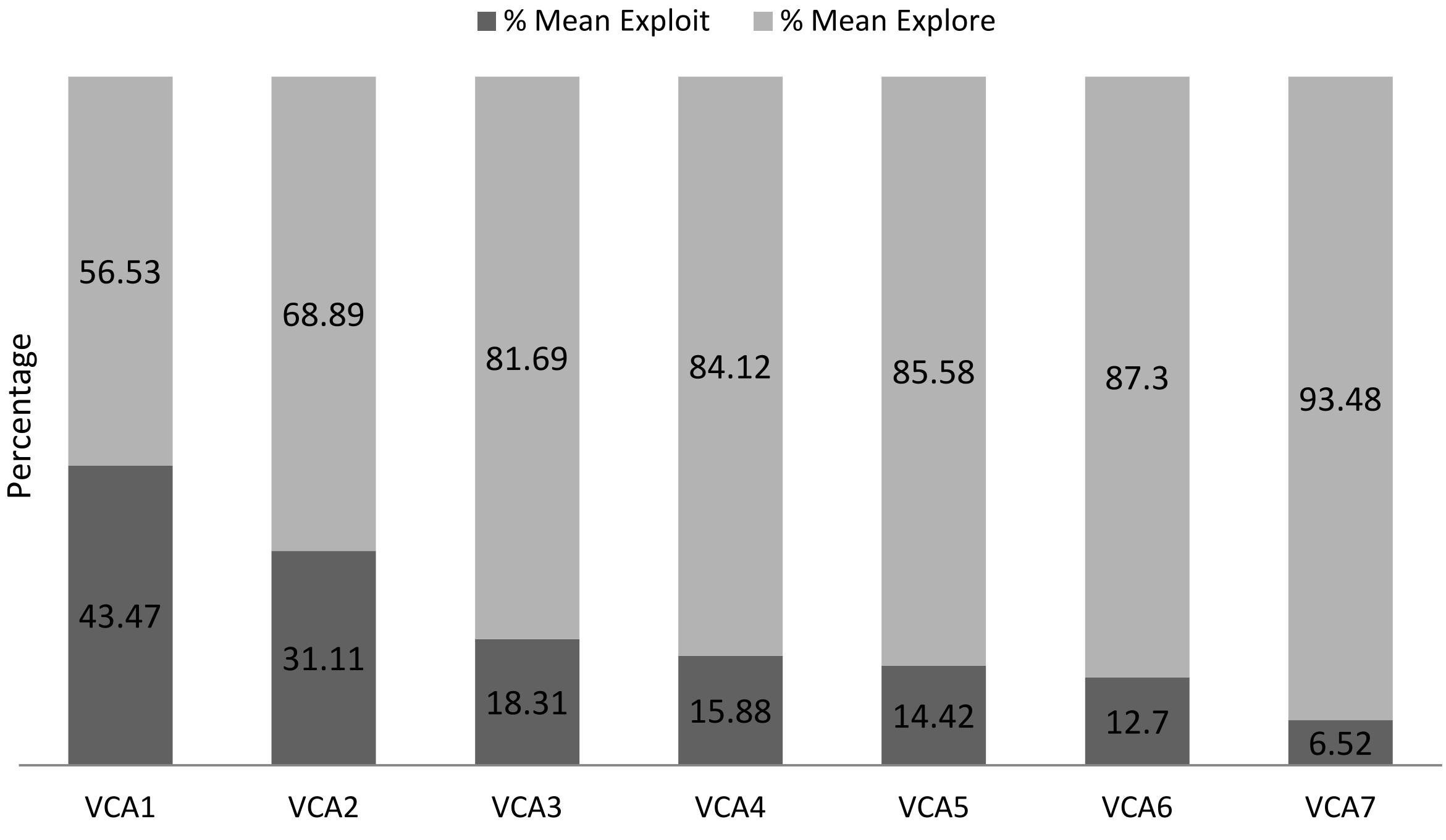

Figure 13. Mean Exploration and Exploitation Percentages of ATLBO for Table 9

## 8. Experimental Observation

Several observations can be elaborated based on the obtained results.

Table 3 shows the size and time performance of the original TLBO and ATLBO. ATLBO outperforms TLBO in three out of six entries in terms of the best test sizes. ATLBO also outperforms TLBO in four out of six entries in terms of the mean test sizes. TLBO outperforms ATLBO in the case of VCA (N; 2, $5^2$ $4^2$ $3^2$, CA (3, $4^2$ $3^2$)) in terms of the mean test size. TLBO and ATLBO have similar execution times for small parameter values for time performances. However, TLBO significantly outperforms ATLBO as the parameter number increases (with fixed $t = 2$) because of the overhead introduced by the fuzzy inference selection. Figure 7 shows that the search gradually favors exploration over exploitation as parameter $p$ increases (with constant $t = 2$ and a small variant of $v$).

Unlike Table 3, Tables 4 to 9 account for the size performance of TLBO and ALTLBO against other meta-heuristic-based strategies. The execution time measures are omitted in this case as the experiments are conducted unfairly based on unequal evaluation of fitness function (e.g., different maximum iteration and control parameters).

ATLBO outperforms all other strategies with the best entries in 17 out of 24 cells in terms of the best test sizes, as shown in Table 4. TLBO and DPSO also provide competitive performance with 14 and 12 best entries, respectively. APSO offers seven best entries, whereas CS provides five best entries. PSTG performs the poorest with only three best entries. ALBO also outperforms the rest of the strategies in terms of the mean test sizes (i.e., with 16 cells). The next closest rival is TLBO (i.e., with five cells) and DPSO (i.e., with four cells). From Figures 8(a) to 8(c), we observe that increasing the parameter value $p$ for the same interaction strength $t$ and values $v$ causes ATLBO to favor exploration. Increasing $t$ similarly causes ATLBO to favor exploration.

DPSO and ATLBO outperform all other strategies with 5 out of 12 cell entries in terms of the test sizes in Table 5. CS is the runner up with 3 cells. APSO provides 2 cells, whereas PSTG offers only 1 cell. DPSO outperforms all other strategies with 5 out of 12 best cell entries in terms of the best mean test size. ATLBO and APSO share 2 best cell entries. PSTG, TLBO, and CS perform the worst with only 1 best cell entry. Referring to Figure 9(a) to 9(c), we observe that increasing the values ($v$) for the same interaction strength ($t$) and parameters ($p$) have small effects in terms of exploration and exploitation. However, increasing $t$ tends to cause ATLBO to increase exploration.

Given the lack of published results, few observations can be made for PSTG, CS, and APSO in Table 6. ATLBO offers the best results in almost all configurations in terms of best test sizes, with the exception of CA ($N$; 2, $5^{10}$) and CA ($N$; 4, $5^{10}$). As the runner up, TLBO (i.e., five out of nine best cell entries) outperforms DPSO (i.e., three out of nine best cell entries). ATLBO outperforms both DPSO and TLBO with four out of nine best entries in terms of the mean test size. Both DPSO and TLBO share the same number of best mean test sizes (i.e., two out of nine cells). The exploration and exploitation of ATLBO in Figures 10(a) to 10(d) show that increasing the values $v$ for the same interaction strength $t$ and parameters $p$ causes a small increase in exploitation. Increasing $t$ tends to cause ATLBO to increase exploration, which is similar to an earlier case.

Table 7 indicates that DPSO and SA outperform all other strategies in terms of the best test size with 8 out of 11 best cell entries. SA can be considered as the best between the two strategies because it offers the best sizes in all participating VCA configurations. Except for DPSO and SA, ATLBO outperforms the others with seven cells. TLBO outperforms PSTG and ACS with five cells. PSTG and ACS perform the worst with four cells. ATLBO outperforms the other strategies with 7 out of 11 best cells in terms of the mean test sizes. Although DPSO has the best results, it has a poorer mean value compared with TLBO with 3 out of 11 cells. PSTG has one entry with the best mean value. Given the lack of published results, information ACS and SA cannot be inferred. Figure 11 shows that increasing sub-configurations {C} tend to increase exploration (with the exception of outlier transitions from VCA6 to VCA7 and from VCA10 to VCA11) with a fixed VCA (N; 2, $3^{15}$, {C}).

Both DPSO and ATLBO outperform all other strategies in Table 8 in terms of the best test size with three out of seven cell entries, followed by TLBO with two cells. HSS and PSTG perform the worst without a single best cell entry. ATLBO outperforms all existing strategies with three out six entries in terms of mean test sizes, followed by DPSO and the TLBO with three out of six and one out of six cell entries, respectively. PSTG and HSS perform the poorest without a single best mean. The chart in Figure 12 shows that increasing sub-configurations {C} also tend to increase exploration similar to the TLBO behavior with a fixed VCA (N; 3, $3^{15}$, {C}).

SA outperforms all other strategies in all VCA configurations in terms of the test size with five entries, as shown in Table 9. DPSO follows with four cell entries, which performs better than ATLBO, TLBO, and HSS (all with three cell entries, respectively). PSTG and ACS have the poorest performance with two and one cell entry, respectively). HSS yields the best results with four out of seven entries for the mean test size. DPSO and ATLBO have the same best entries at three cells. TLBO has only 1 best entry, whereas PSTG has none. Information for ACS and SA cannot be inferred because of the lack of published results. Finally, the chart in Figure 13 shows that our observation is consistent with two earlier findings. In particular, increasing sub-configurations {C} also tend to increase exploration with a fixed VCA (N; 2, $4^3\ 5^3\ 6^2$, {C}).

## 9. Validity Threats

Several validity threats can be associated with our experimental studies. We have identified few threats in this research and elaborated mitigating their effects on our results.

First, the benchmark choice represents an essential threat. We adopt the experimental benchmarks from other well-known studies and experiments in literature. However, we cannot guarantee that these benchmarks represent the actual software configurations in real world. Nevertheless, the benchmarks are derived from configurations of different software programs.

Second, a comparison with other strategies is another threat. Many strategies and tools for generating the $t$-way test suite exist. Given the limited space in this paper and the unavailability of these strategies for implementation within our experimental environment, we cannot compare ALTBO with all other available strategies and tools. To eliminate this threat, Hence, we select the recently published results in a reputable journal for the highest related strategies close to ALTBO (e.g., [20]). In our case, the tuning the parameters of those strategies are out of our control. Nevertheless, our comparison is valid because the published results were obtained with the best tuning parameters.

The original TLBO has twice as much fitness function evaluations as ATLBO, which can also be a significant threat to our experimentations, rendering unfair comparisons. Both the teacher and student phase processes are serially executed per iteration in the original TLBO. In contrast, only one process is selected per iteration in ATLBO based on the adaptive measure of the searching process. Given that both implementations are ours, we can straightforwardly eliminate these threats. We can also ensure that the iteration number within TLBO is always half of that of the ATLBO.

The randomness of the search operators within the meta-heuristic strategies can also be an issue. The best test size results can potentially be obtained at this point by chance and only once out of many runs. Reporting and comparing only the best test size results may not provide a fair indication of the size performance of a particular strategy. Thus, we also relied on the mean results rather than merely focusing on the best test size results.

The choice of fuzzy implementation is another important threat. Utilizing different fuzzy implementation and inference systems may lead to different results (with different membership functions). We recognize that at least two different fuzzy inference system variations exist in literature (e.g., Mamdani-type versus Sugeno-type fuzzy inference systems). Previous studies that adopt fuzzy systems to control different parameters within meta-heuristics apply the Mamdani inference utilizing the center of gravity for the output defuzzification [48]. Most studies often employ either trapezoidal (i.e., as a triangular variant) or Gaussian membership functions. In one such study by Valdez et al [49], it was reported that empirical analysis using both types of membership functions concluded that Trapeziodal membership functions give better performance over Gaussian ones. Hence, we adopt the Mamdani-type fuzzy inference system with a center of gravity and trapezoidal membership functions for our work to obtain a suitable performance.

Finally, the choices of efficiency and performance metrics can also pose as threats. Other metrics that evaluate the efficiency and performance utilizing the internal algorithm structure can exist. However, we adopt the generated size of $t$-way test suites for efficiency and generation time for performance because these metrics are well-known in literature (i.e., for $t$-way test suite construction).

## 10. Concluding Remarks

Results of the comparative experiments show that the usefulness of our approach can be debated further.

We argue that a comparison with the best results is directly influenced by chance (i.e., resulting from the randomness components from each of the meta-heuristics involved). A comparative performance based on the mean alternatively provides a fair indicator of the consistent performance for any particular strategy of interest. Considering both comparisons, ATLBO offers a competitive performance against existing strategies (with DPSO and SA as the closest competitors). Nevertheless, some overhead for ATLBO is observed in terms of the execution time despite its competitive performance. In particular, the processes related to fuzzy inference rules (i.e., calculating the quality measure $Q_m$, intensification measure $I_m$, and diversification measure $D_m$) can be accommodated.

We modify the sequential nature of exploration and exploitation within TLBO. We specifically enhance the original TLBO with adaptive selection; local and global searches are decided at run-time based on the search progress. We maintain the core feature of TLBO (parameter free). Existing meta-heuristics typically adopt specific control parameters and require explicit (problem domain) tuning to ensure a balance between exploration and exploitation. Explicit tuning is unnecessary for ATLBO because the balance between exploration and exploitation is adaptively handled by the implemented fuzzy inference system.

The search pattern of the original TLBO is straightforward, wherein both exploration and exploitation is always at 50%. However, to understand the searching pattern of ATLBO, we need to track the mean percentage of exploration and exploitation taking mixed strength *t*-way test generation problem as the case study. Within the problem of generating mixed *t*-way tests and given VCA (N; $t$, $v^p$, {C}), four main variables of interest exist (i.e., interaction strength *t*, values *v*, parameter *p*, and sub-configuration {C}). We arrive at the following conclusions based on our experiments.

- ATLBO favors exploitation over exploration (i.e., with typical values $p \leq 6$, $t \leq 3$, $v \leq 2$) for small values of *p*, *t*, and *v*.
- When the parameter *p* increases, ATLBO favors exploration over exploitation for a fixed *t* and *v*.
- When the interaction strength *t* increases, ATLBO favors exploration over exploitation for a fixed *p* and *v*.
- When the value *v* increases, ATLBO favors exploration over exploitation for a fixed *p* and *t*. However, the rate of exploration increment is smaller than the effect of the increasing *p* or *v*.
- Given VCA (N; $t$, $v^p$, {C}) and for a fixed *p*, v, and *t*, ATLBO favors exploration over exploitation when {C} increases.

As the search space grows (i.e. horizontally with the increase of *p* and *t*, or vertically with the increase of *v*), ATLBO needs to explore more promising regions to obtain good quality solution. Our findings are consistent with intuition indicating the effectiveness of the developed fuzzy rules.

We are currently looking to adopt ATLBO in other well-known optimization problems (e.g., travelling salesman and vehicle-routing problem) because of its performance. Apart from adopting ATLBO for well-known optimization problems, we are also currently investigating the use of the case-based Reasoning approach in place of our fuzzy inference system to improve its time performance.

Finally, we are also interested in adopting ATLBO to test service-oriented architecture (SOA) solutions. A SOA solution is generally an integrated product set that can be a collection of legacy applications, third-party components, or custom-developed components. Testers are expected to assess not only the individual products and their functionality but also the interaction of components within an overall integrated solution [50, 51]. A strategy such as ATLBO can be useful in assessing the component interaction, which is in line with the continuous growth of SOA with its new features and additional components. In fact, ATLBO can also systematically minimize the test data for testing considerations (i.e., based on the given interaction strength).

## Acknowledgement

This study is funded by the Science Fund Grant for the project entitled, “Constraints T-Way Testing Strategy with Modified Condition/Decision Coverage” from the Ministry of Science, Technology, and Innovation, Malaysia.